\documentclass{ieeeaccess_arxiv}
\usepackage{newunicodechar}
\newunicodechar{ }{\,}

\usepackage{amsmath,amssymb,amsfonts}
\usepackage{algorithmic}
\usepackage{graphicx}
\usepackage{textcomp}
\usepackage{booktabs} 
\usepackage{tabularx}
\usepackage[colorlinks=true, linkcolor=blue, citecolor=blue, urlcolor=blue]{hyperref}
\usepackage[utf8]{inputenc}
\DeclareUnicodeCharacter{200B}{}
\usepackage{array}
\usepackage{bm}
\usepackage{cuted}
\usepackage{caption}
\usepackage{tabularx}
\usepackage{array}
\usepackage{booktabs}
\usepackage{ragged2e}
\usepackage{needspace}
\usepackage{placeins}
\usepackage{adjustbox}
\usepackage{float}
\usepackage{newunicodechar}
\newunicodechar{−}{-}
\usepackage{enumitem}
\usepackage[utf8]{inputenc}
\makeatletter
\AtBeginDocument{\DeclareMathVersion{bold}
\SetSymbolFont{operators}{bold}{T1}{times}{b}{n}
\SetSymbolFont{NewLetters}{bold}{T1}{times}{b}{it}
\SetMathAlphabet{\mathrm}{bold}{T1}{times}{b}{n}
\SetMathAlphabet{\mathit}{bold}{T1}{times}{b}{it}
\SetMathAlphabet{\mathbf}{bold}{T1}{times}{b}{n}
\SetMathAlphabet{\mathtt}{bold}{OT1}{pcr}{b}{n}
\SetSymbolFont{symbols}{bold}{OMS}{cmsy}{b}{n}
\renewcommand\boldmath{\@nomath\boldmath\mathversion{bold}}}
\makeatother 

\def\BibTeX{{\rm B\kern-.05em{\sc i\kern-.025em b}\kern-.08em
    T\kern-.1667em\lower.7ex\hbox{E}\kern-.125emX}}

\begin{document}
\history{}
\doi{}
\title{A Multi-Branch Feature Fusion Approach for Health Misinformation Detection and Propagation}

\author{\uppercase{MKULULI SIKOSANA}\authorrefmark{1}, 
\uppercase{SEAN MAUDSLEY-BARTON}\authorrefmark{1}, and OLUWASEUN AJAO\authorrefmark{1}}

\address[1]{Department Of Computing And Mathematics, Manchester Metropolitan University, Manchester, UK (e-mail: mkululi.sikosana@stu.mmu.ac.uk)}

\tfootnote{This work did not receive any financial support.}

\markboth
{Sikosana \headeretal: Multi-Branch Feature Fusion Approach for Health Misinformation Detection \& Propagation}
{Sikosana \headeretal: Multi-Branch Feature Fusion Approach for Health Misinformation Detection \& Propagation}

\corresp{Corresponding author: MKULULI SIKOSANA (e-mail: mkululi.sikosana@stu.mmu.ac.uk).}

\begin{abstract}
This paper presents a multi-branch fusion framework for detecting and characterising the propagation of health misinformation in online social networks (OSNs). Grounded in the Elaboration Likelihood Model (ELM) and the Theory of Planned Behaviour (TPB), the model fuses transformer-based semantics with rhetorical cues, stance representations, and psychologically motivated proxies in a unified multi-task architecture. In addition to binary classification, we introduce the Cognitive Propagation Score (CPS), an interpretable post-hoc auxiliary score computed from psychologically motivated, text-derived cues capturing argument complexity, emotional intensity, and content-derived virality potential, to support diffusion-risk reasoning when engagement ground truth is incomplete or unavailable. Experiments on three benchmark datasets, Constraint, COVID--19\_FNIR, and Monkeypox, show strong classification performance, achieving ROC--AUC up to 0.9999 on COVID--19\_FNIR, while propagation-oriented ranking achieves near-perfect agreement when engagement-derived supervision is available (Monkeypox, Spearman's $\rho = 0.9952$) and similarly high ranking alignment under proxy-based supervision on COVID--19\_FNIR ($\rho = 0.9954$). Compared with representative literature baselines, the fusion model improves detection on Constraint and COVID--19\_FNIR, while Monkeypox remains more challenging, reflecting domain- and signal-specific differences. Ablation analysis further indicates that psychological and rhetorical branches provide complementary gains beyond semantic embeddings. Overall, the framework bridges cognitive theory and neural modelling to improve transparency and to support scalable misinformation monitoring, with future work required to validate CPS against human-centred diffusion judgements.
\end{abstract}


\titlepgskip=-21pt

\maketitle

\section{Introduction}
\label{sec:introduction}
\PARstart{H}{ealth} misinformation on online social networks (OSNs) threatens public health, institutional trust, and policy-making \cite{lazer2018science,vosoughi2018spread}. During COVID-19, misleading claims circulated at incredible speed, contributing to unsafe behaviours and eroding confidence in health guidance \cite{gallotti2020assessing,cinelli2020covid,islam2020covid,allcott2017socialmedia}. While manual fact-checking is essential, it does not scale to the volume and speed of modern infodemics \cite{patwa2021fighting}.

Transformer-based NLP models, including BERT \cite{devlin2019bert} and RoBERTa \cite{liu2019roberta}, have improved veracity classification by capturing contextual language patterns \cite{wani2021evaluating}. However, many high-performing detectors offer limited behavioural insight into \emph{why} content persuades and diffuses. Persuasion theories provide a useful interpretive lens. The Elaboration Likelihood Model (ELM) distinguishes between central-route processing (argument quality and cognition) and peripheral-route processing (heuristics and affect) \cite{petty1986elaboration}. The Theory of Planned Behaviour (TPB) explains behavioural intention via attitudes, subjective norms, and perceived behavioural control \cite{ajzen1991theory}. These constructs motivate feature design that is more transparent than token-level attention alone.

A practical barrier for propagation modelling is that engagement traces are not uniformly available across datasets. Some corpora include engagement metadata, whereas others provide only veracity labels. This motivates diffusion-oriented signals that can be computed consistently from text, enabling ranking and prioritisation even when native engagement supervision is missing.

This paper proposes a cognitively grounded, multi-branch, multi-task framework for misinformation detection and diffusion-oriented analysis. The model fuses transformer embeddings with stance, rhetorical structure, and ELM and TPB proxies extracted from the post text. In addition, this paper also introduces the \textbf{Cognitive Propagation Score (CPS)}, a theory-aligned post-hoc auxiliary score designed to support interpretability by mapping psychologically meaningful cues to a single diffusion-propensity scalar (that is, a content-derived indicator of relative likelihood of rapid spread or “virality”, not a retweet count). CPS is constructed to be engagement-independent, since CPS subfeatures exclude observed engagement fields. Observed engagement, when available, is used for propagation-regression supervision and auxiliary comparison, but it does not enter the CPS calculation.

In practice, this means the engineered branches operationalise theory as observable language cues, where TPB-aligned proxies reflect Attitude (for example affective polarity and certainty), Subjective Norms (for example inclusive pronouns and normative phrasing), and Perceived Behavioural Control (for example directive or instructional language), while ELM-aligned proxies capture central-route cues (for example readability and lexical diversity) and peripheral-route cues (for example emphatic punctuation, capitalisation, and urgency markers).

\textbf{The study makes the following contributions:}
\begin{enumerate}
    \item Introduce a multi-branch fusion architecture that combines semantic, rhetorical, stance, and behavioural theory-informed features for health misinformation detection.
    \item Operationalise ELM and TPB constructs into explicit, text-derived proxy features that support feature-level interpretation.
    \item Propose CPS as an interpretable auxiliary diffusion-propensity signal, intended for prioritisation when engagement ground truth is unavailable or delayed.
    \item Evaluate the framework on three health-related benchmarks (Constraint, COVID-19-FNIR, and Monkeypox datasets) and quantify the contribution of each feature branch via ablation analysis.
\end{enumerate}

In doing so, the study lays the foundation for more effective and interpretable tools in combating health misinformation online.

\section{Related Work and Benchmark Baselines}
\label{sec:baseline_models_for_Benchmark_dataset}

\subsection{Baseline rationale and comparability boundary}
To situate this study within the health-misinformation literature, baseline model families most commonly reported for each benchmark dataset, spanning classical machine learning, transformer-based classifiers, and lightweight neural architectures, are summarised. The baseline results discussed in this section are taken from the original studies, and they are therefore interpreted as \emph{study-specific} because prior work frequently differs in preprocessing, label definitions, data splits (hold-out versus $k$-fold cross-validation), and hyperparameter tuning procedures. Accordingly, these reported baselines are used to characterise the evaluation landscape and to motivate the methodological gaps addressed by our experiments, rather than as a strictly like-for-like numerical benchmark across papers.

\subsection{Baseline Models on the Constraint COVID-19 Fake News Dataset (English)}
This section summarises reported baselines from the literature for each benchmark dataset. Comparability and fairness, including the boundary between our implemented experiments and literature-reported baselines used for contextual reference, is handled in Section~\ref{subsec:baseline_fairness}.

Constraint@AAAI-2021, a shared task dataset comprising 10,700 social media posts labelled as real or fake, has established several strong baselines \cite{patwa2021fighting}. The top-performing model in the original competition combined XLNet with topic-distribution features extracted via Latent Dirichlet Allocation (LDA), achieving an F1-score of approximately 96.7\% \cite{gautam2021fake}. This XLNet+LDA fusion outperformed vanilla transformer models, demonstrating the added value of topic-level cues.

As a non-neural baseline, a linear Support Vector Machine (SVM) trained on handcrafted linguistic features (n-grams, readability metrics, punctuation, etc.) attained a weighted F1-score of about 95.2\% on the test set \cite{felber2021machine}. Among traditional machine learning approaches, this SVM performed best, surpassing alternatives such as random forests and Naïve Bayes \cite{ahmed2022explainable}. Notably, the dataset creators themselves reported SVM to be the most effective among four classical algorithms, with an F1-score of 93.4\% \cite{patwa2021fighting}.

   Additional transformer-based studies on the Constraint dataset included CT-BERT ensembling by \cite{glazkova2021g2tmn}, pretrained language model ensembling by \cite{li2021exploring}, and broader transformer comparisons by \cite{sadiq2021transformer} and \cite{alghamdi2023towards}. These studies confirm the strong performance of transformer approaches on the dataset and provide useful context beyond the original shared-task baselines. For comparative purposes in this paper, the XLNet+LDA model (\textasciitilde96.7\% F1) and the linguistic-feature SVM (\textasciitilde95\% F1) are treated as the main reference baselines for the Constraint dataset.

\subsection{Baselines on the COVID--19\_FNIR Dataset}

The COVID--19\_FNIR dataset contains approximately 7,600 labelled social media posts concerning COVID-19 misinformation \cite{saenz2021covid}. The dataset creators reported that a fine-tuned DistilBERT model, trained without auxiliary features or domain-specific pretraining, achieved an F1-score of approximately 93\% \cite{qadees2023cross}. This result has been corroborated in subsequent studies. For example, in previous work co-authored by the first author, Sikosana et al.\ \cite{sikosana2024hybrid} found that DistilBERT classifiers attained F1-scores as high as 97\%, while CNN--LSTM models using standard word embeddings achieved accuracy values of 98--99\%.

Traditional machine learning approaches also performed competitively on COVID--19\_FNIR. Random forest and SVM classifiers reached 91–94\% F1, suggesting that the dataset contains strong linguistic cues that can be learned by both neural and classical models \cite{qadees2023cross}. In comparative studies, a linear SVM attained 94.4\% F1, and a tuned BiLSTM model with Word2Vec embeddings reached 99\%, outperforming DistilBERT and RoBERTa baselines \cite{qadees2023cross}. Across the literature, reported performance varies by split strategy and experimental protocol. These studies are used to motivate the set of representative baselines evaluated under the unified protocol (Section~\ref{subsec:baseline_fairness}).

Further studies have explored explainable transformer-based approaches, such as DistilBERT integrated with SHAP explanations \cite{ayoub2021combat}. However, these models prioritised interpretability over performance improvement. Traditional classifiers, including logistic regression, were also evaluated but consistently underperformed relative to transformer architectures. Based on these findings, fine-tuned DistilBERT, achieving approximately 93\% F1, is adopted as the baseline benchmark for the COVID--19\_FNIR dataset.

\subsection{Baselines on the Monkeypox Dataset}

Misinformation detection in the mpox domain is a relatively recent field. PoxVerifi introduced a curated dataset of 225 mpox-related claims (170 true, 55 false) and trained a BERT-based classifier for claim-level misinformation detection \cite{kolluri2022poxverifi}. On this claim-level benchmark, the model achieved approximately 96\% accuracy under 10-fold cross-validation. Labels were derived from trusted public health sources, including the WHO, and no external data or domain adaptation was used.

PoxVerifi~\cite{kolluri2022poxverifi} provides contextual evidence that transformer encoders can separate mpox misinformation under curated, claim-level conditions. Comparability and fairness, including which baselines are treated as contextual literature references rather than directly comparable benchmarks under our unified protocol, are specified in Section~\ref{subsec:baseline_fairness}.

\subsection{Summary of Baseline Coverage}

Across the three datasets, the literature provides benchmarks from both conventional and neural model families. For the Constraint dataset, one of the strongest benchmarks reviewed in this paper is approximately 96.7\% $F_1$ for an XLNet+LDA model, while linguistically informed SVM variants have achieved approximately 95\% $F_1$ \cite{gautam2021fake,felber2021machine}. For COVID--19\_FNIR, reported results range from approximately 92--93\% $F_1$ for DistilBERT-based classification \cite{qadees2023cross} to approximately 98--99\% for CNN--LSTM hybrid models reported in previous work co-authored by the first author \cite{sikosana2024hybrid}. These findings indicate that performance can approach ceiling levels on this corpus under particular data splits, preprocessing procedures, and evaluation protocols. For mpox, systems such as PoxVerifi \cite{kolluri2022predictors} provide useful context for claim-level verification but are not directly comparable with the tweet-level \textit{Monkeypox misinformation dataset} used in this paper. PoxVerifi is therefore treated as a contextual reference, while tweet-level baselines are prioritised for like-for-like performance comparisons.

Overall, these values provide contextual reference points and literature-reported upper bounds. In this paper, dataset-level state-of-the-art claims based solely on cross-paper comparisons where splits, preprocessing, or validation regimes differ are avoided. Instead, emphasis is on the incremental value of feature fusion and psychologically grounded modelling under a fixed, reproducible protocol, with Table~\ref{tab:baseline_comparison} reporting the baselines used for direct comparison and Section~\ref{sec:baseline_comparison} stating the comparability boundary explicitly.

\section{Materials \& Methods}
\subsection{Dataset and Preprocessing}
This study uses three misinformation datasets: COVID--19\_FNIR \cite{saenz2021covid}, Constraint \cite{patwa2021fighting}, and the \textit{Monkeypox dataset} \cite{crone2022monkeypox}. The datasets vary in size and annotation richness. To preserve methodological consistency across
experiments, all three datasets were processed using a common preprocessing and feature-engineering pipeline. This included tokenisation, lowercasing, stop-word removal, and punctuation standardisation, together with the extraction and scaling of the engineered feature sets used throughout this study. COVID--19\_FNIR and Constraint are binary-labelled for fake versus
real news classification. The Monkeypox dataset includes engagement metadata, including likes,
retweets, and replies. These variables are not used as input features
to the misinformation-detection model or in the construction of CPS.
Instead, they are used to construct the observed engagement target for
propagation-regression supervision and for auxiliary evaluation. CPS is computed separately from the engagement-independent, text-derived quantities specified in Eq.~\eqref{eq:cps-raw}.

\subsection{Feature Extraction}
This section details the computational feature extraction process used to derive meaningful behavioural, semantic, and rhetorical representations from the input tweet text. The extracted features are grouped into five categories: Transformer Embeddings, Rhetorical Features, Stance Features, ELM Features, and TPB Features. The scaled feature representations form the inputs to the classification and propagation-regression components, while the raw text-derived quantities required by Eq.~\eqref{eq:cps-raw} are retained separately
for post-hoc CPS computation.

\begin{itemize}
    \item \textbf{Semantic (text) branch:} \(B(x_i)=f_{\mathrm{DistilBERT}}(x_i) \in\mathbb{R}^{768}\).
    A pretrained DistilBERT model encodes post \(x_i\) as a     768-dimensional dense contextual representation \(B(x_i)\). This representation captures semantic patterns beyond simple word
    frequency and constitutes the semantic branch of the fusion architecture. The terms \textit{semantic branch}, \textit{text branch}, and \textit{textual branch} in earlier descriptions all refer to this same DistilBERT-based component. To avoid ambiguity, it is referred to consistently hereafter as
    the \textit{semantic (text) branch}.

    \item \textbf{Rhetorical Features:} $R = f_{\text{rhetoric}}(x), \quad R \in \mathbb{R}^{d_R}$ \\
    These features capture discursive and stylistic cues (e.g., clickbait indicators, argument complexity, writing style) that distinguish credible from sensational content.

    \item \textbf{Stance Features:} $S = f_{\text{stance}}(x), \quad S \in \mathbb{R}^{d_S}$ \\
    This representation encodes the attitudinal stance of the author (support, denial, query, comment) towards a claim.

    \item \textbf{Psychological Features (ELM \& TPB):} $[E;T] = [f_{\text{ELM}}(x); f_{\text{TPB}}(x)], \quad E \in \mathbb{R}^{d_E},\ T \in \mathbb{R}^{d_T}$ \\
    Features derived from ELM and TPB capture behavioural and persuasive cues, including central/peripheral signals and perceived behavioural control.
\end{itemize}

For dimensional consistency, \(B(x_i)\in\mathbb{R}^{768}\), \(R(x_i)\in\mathbb{R}^{4}\), \(S(x_i)\in\mathbb{R}^{4}\), \(E(x_i)\in\mathbb{R}^{10}\), and \(T(x_i)\in\mathbb{R}^{9}\), matching the fused input definition in Section~\ref{sec:fusion-hidden}.

\subsection{ELM Feature Operationalisation}
Each tweet $x_i \in X \subset \mathbb{R}^n$ is mapped to a psychologically interpretable feature vector $E(x_i) \in \mathbb{R}^d$, composed of central and peripheral route cues. These are computed using standard NLP libraries (e.g., spaCy, TextBlob, VADER), as follows:

\noindent\textit{Proxy note.} These features are linguistic proxies intended to reflect theoretically motivated cues, rather than direct measurements of elaboration, beliefs, or psychological processing states.

\subsubsection*{Central Route Features}

The ELM central-route representation comprises five interpretable
text-derived features:

\[
C(x_i)=
[c_1(x_i),\,c_2(x_i),\,c_3(x_i),\,c_4(x_i),\,c_5(x_i)]
\in \mathbb{R}^{5}.
\]

\setcounter{equation}{0}

\noindent\textit{Flesch--Kincaid Grade Level (FKGL).}
The first feature measures textual readability and is defined as

\begin{equation}
\begin{aligned}
c_1(x_i)
&=
0.39
\left(
\frac{\text{words}(x_i)}
{\text{sentences}(x_i)}
\right)\\
&\quad+
11.8
\left(
\frac{\text{syllables}(x_i)}
{\text{words}(x_i)}
\right)
-15.59 .
\end{aligned}
\label{eq:fkgl}
\end{equation}

\noindent\textit{Type--Token Ratio (TTR).}
Lexical diversity is represented by the proportion of unique tokens
relative to the total number of tokens:

\begin{equation}
c_2(x_i)
=
\frac{
|\text{UniqueTokens}(x_i)|
}{
|\text{Tokens}(x_i)|
}.
\label{eq:ttr}
\end{equation}

\noindent\textit{Mean sentiment polarity.}
The mean polarity of the words in post \(x_i\) is calculated as

\begin{equation}
c_3(x_i)
=
\frac{1}{|x_i|}
\sum_{w_j\in x_i}
\text{Polarity}(w_j).
\label{eq:mean-polarity}
\end{equation}

\noindent\textit{Token count.}
Post length is represented by the total number of tokens:

\begin{equation}
c_4(x_i)=|x_i|.
\label{eq:token-count}
\end{equation}

\noindent\textit{Average sentence length.}
Average sentence length is calculated as the ratio of the total number
of words to the total number of sentences:

\begin{equation}
c_5(x_i)
=
\frac{\text{Total Words}}
{\text{Total Sentences}}.
\label{eq:avg-sentence-length}
\end{equation}

\paragraph{TTR calculation and scaling.}
The raw TTR defined in Eq.~\eqref{eq:ttr} is naturally bounded between
0 and 1. Before model training and feature fusion, TTR was min--max
scaled together with the other engineered rhetorical, stance, ELM,
and TPB features. The scaling parameters were estimated from the
training partition only:

\[
\widetilde{c}_2(x_i)
=
\frac{
c_2(x_i)-c_{2,\min}^{\mathrm{train}}
}{
c_{2,\max}^{\mathrm{train}}
-
c_{2,\min}^{\mathrm{train}}
}.
\]

The same training-derived minimum and maximum were applied unchanged
to the validation and test partitions. This prevented information from
the validation or test data from influencing feature scaling. Thus,
TTR was min--max normalised rather than \(z\)-standardised.

Min--max scaling places TTR on a comparable numerical range with the
other engineered features but does not remove its known dependence on
text length. Shorter posts can therefore produce higher or more
variable TTR values. A moving-average or fixed-window lexical-diversity
measure, such as MATTR or MSTTR, was not applied. TTR is consequently
interpreted as an approximate within-corpus indicator of lexical
diversity rather than a length-independent measure.

\subsubsection*{Peripheral Route Features}

The ELM peripheral-route representation comprises five interpretable
text-derived features capturing emotional, stylistic, and heuristic
appeal:

\[
P(x_i)=
[p_1(x_i),\,p_2(x_i),\,p_3(x_i),\,p_4(x_i),\,p_5(x_i)]
\in \mathbb{R}^{5}.
\]

\noindent\textit{Peripheral feature vector.}
The complete peripheral-route representation is defined as

\begin{equation}
P(x_i)=
[p_1(x_i),\,p_2(x_i),\,p_3(x_i),\,p_4(x_i),\,p_5(x_i)]
\in \mathbb{R}^{5}.
\tag{6}
\label{eq:peripheral-vector}
\end{equation}

\noindent\textit{Exclamation-mark ratio.}
Emphatic punctuation is represented by the proportion of exclamation
marks relative to post length:

\begin{equation}
p_1(x_i)
=
\frac{
|\texttt{``!'' in }x_i|
}{
|x_i|
}.
\tag{7}
\label{eq:exclamation-ratio}
\end{equation}

\noindent\textit{Question-mark ratio.}
Questioning or interrogative punctuation is represented by the
proportion of question marks relative to post length:

\begin{equation}
p_2(x_i)
=
\frac{
|\texttt{``?'' in }x_i|
}{
|x_i|
}.
\tag{8}
\label{eq:question-ratio}
\end{equation}

\noindent\textit{Uppercase-letter ratio.}
Capitalisation intensity is represented by the proportion of uppercase
letters among all alphabetic characters:

\begin{equation}
p_3(x_i)
=
\frac{
|\text{UppercaseLetters}(x_i)|
}{
|\text{AllLetters}(x_i)|
}.
\tag{9}
\label{eq:uppercase-ratio}
\end{equation}

\noindent\textit{All-uppercase word count.}
The use of ``shouting'' language is represented by the number of words
containing at least two characters that are written entirely in
uppercase:

\begin{equation}
p_4(x_i)
=
\left|
\left\{
w_j \in x_i :
w_j=\text{uppercase}(w_j)
\land |w_j|\geq 2
\right\}
\right|.
\tag{10}
\label{eq:allcaps-count}
\end{equation}

\noindent\textit{Urgency-term ratio.}
Urgency is represented by the proportion of terms drawn from the
predefined urgency lexicon
\{\texttt{now}, \texttt{urgent}, \texttt{act},
\texttt{immediately}\}:

\begin{equation}
p_5(x_i)
=
\frac{
|\text{UrgentTerms}(x_i)|
}{
|x_i|
}.
\tag{11}
\label{eq:urgency-ratio}
\end{equation}

\noindent
Together, the central- and peripheral-route features form the complete
ELM representation:

\[
E(x_i)
=
[C(x_i),\,P(x_i)]
\in \mathbb{R}^{10}.
\]

\subsection{TPB Feature Operationalisation}
The TPB posits that behavioural intention (e.g., sharing misinformation) is influenced by three psychological constructs: Attitude (A), Subjective Norms (SN), and Perceived Behavioural Control (PBC). In this study, these components are extracted using natural language processing heuristics:

\noindent\textit{Proxy note.} Because TPB constructs are classically measured via self-report, the features below should be interpreted as scalable text-derived proxies that approximate attitude, norms, and control cues, not as definitive evidence of a user’s internal intention.

\begin{itemize}
    \item \textbf{Attitude-aligned textual tone (\(A\))} is approximated using mean sentiment polarity, lexical valence, and modality. These features capture the evaluative tone and degree of commitment expressed towards the topic or claim presented in the post. They do not directly measure the author's attitude towards misinformation as a general category or their attitude towards the behaviour of sharing misinformation.
    \item \textbf{Subjective Norms (SN)} are inferred from group pronouns (e.g., ``we'', ``us''), retweet or mention indicators (e.g., ``@user'', ``RT''), and social comparison cues (e.g., ``everyone'', ``others''). Because engagement metadata is unavailable for Constraint and COVID--19\_FNIR, subjective-norm proxies are computed from textual markers rather than interaction counts. Where a specific marker is absent in a post, the corresponding feature is set to zero after preprocessing, preserving a consistent 9-dimensional TPB vector across datasets.\par
    
   \item \textbf{Perceived Behavioural Control (PBC)} is captured through epistemic certainty cues (e.g., ``definitely'', ``surely''), instructional language (e.g., ``share this'', ``click''), and hashtag imperatives (e.g., \#WakeUp, \#JoinUs).
\end{itemize}

Based on these operationalisations, a 9-dimensional TPB-aligned feature vector for each tweet is constructed. Each component $t_k(x_i)$ corresponds to a specific linguistic proxy representing one of the TPB constructs: Attitude ($t_1$ to $t_3$), Subjective Norms ($t_4$ to $t_6$), and Perceived Behavioural Control ($t_7$ to $t_9$). The resulting feature vector is:

\[
T(x_i) = [t_1(x_i), t_2(x_i), \dots, t_9(x_i)] \in \mathbb{R}^9
\]

\noindent\textit{Grouped view (for interpretability).}
For interpretability, the nine TPB-aligned features are grouped
according to the three theoretical constructs of Attitude,
Subjective Norms, and Perceived Behavioural Control:

\[
\begin{aligned}
T(x_i)
&=
[A(x_i),\,SN(x_i),\,PBC(x_i)],\\
A(x_i)
&=
[t_1(x_i),\,t_2(x_i),\,t_3(x_i)],\\
SN(x_i)
&=
[t_4(x_i),\,t_5(x_i),\,t_6(x_i)],\\
PBC(x_i)
&=
[t_7(x_i),\,t_8(x_i),\,t_9(x_i)].
\end{aligned}
\]

\paragraph{TPB measurement boundary.}
In the canonical TPB, attitude refers to an individual's favourable or
unfavourable evaluation of a specified behaviour. For the present study,
the relevant behaviour is sharing or engaging with a health-information
post. The datasets do not contain direct self-report measures of
attitudes, subjective norms, perceived behavioural control, or
behavioural intentions. Consequently, \(A(x_i)\), \(SN(x_i)\), and
\(PBC(x_i)\) are theory-aligned, text-derived proxies rather than direct
measurements of the corresponding latent psychological constructs.

\subsubsection*{Attitude-Aligned Textual Tone Proxy}

The attitude-aligned component comprises three text-derived features
representing affective valence and strength of expressed commitment.

\setcounter{equation}{11}

\noindent\textit{Mean sentiment polarity.}
The mean sentiment polarity expressed in post \(x_i\) is calculated as

\begin{equation}
t_1(x_i)
=
\frac{1}{|x_i|}
\sum_{w_j \in x_i}
\text{Sentiment}(w_j).
\label{eq:tpb-sentiment}
\end{equation}

\noindent\textit{Lexical valence score.}
Affective and evaluative word use is represented by a lexical valence
score derived from an opinion lexicon:

\begin{equation}
t_2(x_i)
=
\text{Lexical Valence Score}(x_i).
\label{eq:tpb-valence}
\end{equation}

\noindent\textit{Modality ratio.}
The strength or qualification of the expressed position is represented
by the proportion of modal verbs among all verbs:

\begin{equation}
t_3(x_i)
=
\frac{
|\text{ModalVerbs}(x_i)|
}{
|\text{Verbs}(x_i)|
}.
\label{eq:tpb-modality}
\end{equation}

Here, \(t_1\) and \(t_2\) represent the affective and evaluative valence
expressed towards the post's topic or claim, whereas \(t_3\) represents
the strength or qualification of the expressed position through
modality. In a conventional TPB study, attitude would be measured in
relation to a clearly specified behaviour, which in this study would be
sharing or engaging with a health-information post. Because the datasets
do not contain survey measures of users' behavioural evaluations or
intentions, these text-derived variables are treated only as
attitude-aligned proxies. They should not be interpreted as direct
measurements of an author's attitude towards misinformation or their
intention to share it.

\paragraph{VADER validity and observed limitations.}
VADER was originally developed and evaluated specifically for
social-media text. Hutto and Gilbert~\cite{hutto2014vader} reported a
correlation of \(r=0.881\) between VADER sentiment-intensity scores and
aggregated human judgements of tweets. For three-class classification
into positive, neutral, and negative sentiment, VADER achieved an
\(F_1\)-score of 0.96, compared with 0.84 for individual human raters.
The correlation is the more relevant benchmark here because this study
uses the continuous VADER compound score rather than treating VADER
output as a gold-standard sentiment class.

Corpus-specific accuracy could not be calculated because the datasets do
not contain independently validated sentiment annotations. Qualitative
inspection nevertheless identified posts for which lexicon-based
sentiment should be interpreted cautiously. These included posts
containing sarcasm, quotations, implicit evaluations, domain-specific
health terminology, and reports of emotionally negative events written
in otherwise neutral or advisory language. Quoted or reported claims
presented a particular difficulty because VADER scores the sentiment
expressed in the words but cannot reliably determine whether the author
endorses, rejects, or merely reports the claim. Negation scope and mixed
sentiment within a single post also produced potentially misleading
compound scores. Accordingly, VADER is used as an approximate indicator
of expressed textual valence and not as a definitive measure of the
author's attitude, emotional state, or intention to share misinformation.

\subsubsection*{Subjective Norms}

The subjective-norm component comprises three text-derived proxies for
group identification, social referencing, and perceived normative cues.

\noindent\textit{Group-pronoun ratio.}
Group identification is represented by the proportion of group-based
pronouns, such as ``we'' and ``us'', among all pronouns:

\begin{equation}
t_4(x_i)
=
\frac{
|\text{GroupPronouns}(x_i)|
}{
|\text{AllPronouns}(x_i)|
}.
\label{eq:tpb-group-pronouns}
\end{equation}

\noindent\textit{Retweet/mention token count.}
Textual indicators of social referencing are represented by the number
of \texttt{@mention} and \texttt{RT} tokens present in the post:

\begin{equation}
t_5(x_i)
=
\text{Retweet/Mention Count}(x_i).
\label{eq:tpb-mentions}
\end{equation}

\noindent\textit{Social-comparison cue count.}
Normative or comparative language is represented by the number of social
comparison expressions, such as ``everyone'', ``others'', and ``no one'':

\begin{equation}
t_6(x_i)
=
\text{Social Comparison Cue Count}(x_i).
\label{eq:tpb-social-comparison}
\end{equation}

The features \(t_4\), \(t_5\), and \(t_6\) are textual proxies for
subjective-norm cues rather than measurements of actual peer behaviour
or perceived social pressure. Where a particular marker is absent, its
corresponding feature value is set to zero.

\subsubsection*{Perceived Behavioural Control}

The perceived-behavioural-control component comprises three proxies for
certainty, instructional language, and directive framing.

\noindent\textit{Certainty-cue ratio.}
Epistemic certainty is represented by the proportion of certainty cues,
such as ``definitely'' and ``surely'', relative to post length:

\begin{equation}
t_7(x_i)
=
\frac{
|\text{CertaintyCues}(x_i)|
}{
|x_i|
}.
\label{eq:tpb-certainty}
\end{equation}

\noindent\textit{Instructional-language count.}
Behaviourally directive language, including expressions such as
``click here'', ``check this'', and ``share this'', is represented by

\begin{equation}
t_8(x_i)
=
\text{Instructional Language Count}(x_i).
\label{eq:tpb-instruction}
\end{equation}

\noindent\textit{Hashtag-directive count.}
Imperative or mobilisation-oriented hashtags, such as
\texttt{\#WakeUp} and \texttt{\#JoinUs}, are represented by

\begin{equation}
t_9(x_i)
=
\text{Hashtag Directive Count}(x_i).
\label{eq:tpb-hashtags}
\end{equation}

Together, the nine features form the complete TPB-aligned representation

\[
T(x_i)
=
[t_1(x_i),\,t_2(x_i),\,\ldots,\,t_9(x_i)]
\in \mathbb{R}^{9}.
\]

For neural fusion, all engineered ELM and TPB features are min--max
scaled to \([0,1]\) using parameters estimated from the training
partition. Separately, the raw quantities required for CPS, namely
Flesch--Kincaid Grade Level, adjective count, VADER compound sentiment,
punctuation count, and character length, are retained for the post-hoc
calculation defined in Eq.~\eqref{eq:cps-raw}.

\subsection{Theoretical Justification for CPS and TPB Operationalisation}

Traditional applications of the TPB rely on self-report instruments to measure latent constructs such as attitude, subjective norms, and perceived behavioural control. While this approach offers theoretical robustness, it lacks scalability and cannot be applied retroactively to large-scale social media corpora.

In this study, a computational social science approach is adopted to
reformulate ELM- and TPB-related constructs as psycholinguistic proxies
for the learned fusion architecture. These features provide the
theoretical basis for the psychological branch of the multi-branch
model, but the complete ELM and TPB feature vectors should not be
interpreted as direct components of CPS.

The post-hoc CPS follows the original scoring implementation described
in Eq.~\eqref{eq:cps-raw}. It uses a narrower set of text-derived
quantities comprising Flesch--Kincaid Grade Level, an adjective-count
emotional-intensity proxy, VADER compound sentiment, punctuation count,
and character length. Accordingly, ELM and TPB provide the broader
theoretical framing of the fusion architecture, whereas CPS is a
separate, engagement-independent, text-derived auxiliary index.
Propagation regression remains a distinct learned task supervised by
observed engagement or the transferred propensity-to-spread proxy,
depending on dataset availability.

\subsection{Multi-Branch Feature Fusion \& Hidden Representation}
\label{sec:fusion-hidden}
\setcounter{equation}{20}

Each input tweet is transformed into multiple streams:
\begin{itemize}
  \item Semantic (text) branch: \(B(x_i)=f_{\text{DistilBERT}}(x_i)\in\mathbb{R}^{768}\).
This branch encodes the post text using the 768-dimensional DistilBERT contextual representation.
  \item Rhetorical branch: \( R(x_i) \in \mathbb{R}^4 \)
  \item Stance branch: \( S(x_i) \in \mathbb{R}^4 \)
  \item ELM branch: \( E(x_i) \in \mathbb{R}^{10} \)
  \item TPB branch: \( T(x_i) \in \mathbb{R}^9 \)
\end{itemize}

\noindent\textit{Terminology note.}
The semantic branch and the text branch refer to the same model component, namely the DistilBERT representation \(B(x_i)\). To avoid ambiguity, the term \textit{semantic (text) branch} is used throughout
the remainder of this paper, including the ablation analysis.

The final fused representation is defined as:
\begin{equation}
F(x_i) = \text{Concat}(B(x_i), R(x_i), S(x_i), E(x_i), T(x_i)) \in \mathbb{R}^{d_{\text{fused}}}
\end{equation}

This vector aggregates five feature branches: semantic (text) embeddings \(B(x_i)\), rhetorical cues \(R(x_i)\), stance representations \(S(x_i)\), ELM features \(E(x_i)\), and TPB features \(T(x_i)\).

\paragraph{}
Table~\ref{tab:feature-inventory} provides the reference definition of all engineered features used in this paper, including names, operational definitions, dimensionality, and dataset availability, and it also serves as the reference definition for the fused input representation used throughout this paper.

\begin{strip}
\centering

\captionof{table}{Feature inventory and branch dimensionality used in the fusion model.}
\label{tab:feature-inventory}

\vspace{2pt}

\small
\setlength{\tabcolsep}{4.5pt}
\renewcommand{\arraystretch}{1.12}

\begin{tabularx}{\textwidth}{
    >{\raggedright\arraybackslash}p{2.4cm}
    >{\centering\arraybackslash}p{1.4cm}
    >{\centering\arraybackslash}p{0.9cm}
    >{\raggedright\arraybackslash}X
    >{\raggedright\arraybackslash}p{2.8cm}
}

\toprule

\textbf{Branch} &
\textbf{Symbol} &
\textbf{Dim.} &
\textbf{Features (names and definitions)} &
\textbf{Availability} \\

\midrule

Semantic (DistilBERT) &
\(B(x_i)\) &
768 &
Contextual embedding from DistilBERT-base-uncased for tweet \(x_i\). &
All datasets \\[2pt]

Rhetorical &
\(R(x_i)\) &
4 &
\(r_1\): discourse-cue density, defined as the count of discourse connectives normalised by tokens;

\(r_2\): attribution-marker density, including cues such as ``according to'', ``source'', citations, and URLs, normalised by tokens;

\(r_3\): elaboration-marker density, including causal and contrast connectives such as ``because'' and ``however'', normalised by tokens;

\(r_4\): clickbait/sensational cue index, defined as the normalised proportion of sensational terms. &
All datasets (text-derived) \\[2pt]

Stance &
\(S(x_i)\) &
4 &
Four-way stance representation over \{support, denial, query, comment\}, encoded as a one-hot vector or predicted class probabilities where applicable. &
All datasets (text-derived) \\[2pt]

ELM &
\(E(x_i)\) &
10 &
\textit{Central route:}
\(c_1\) FKGL;
\(c_2\) TTR, defined as the unique-to-total token ratio;
\(c_3\) mean polarity;
\(c_4\) token count;
\(c_5\) average sentence length.

\textit{Peripheral route:}
\(p_1\) exclamation ratio;
\(p_2\) question ratio;
\(p_3\) uppercase ratio;
\(p_4\) all-caps count;
\(p_5\) urgency-term ratio. &
All datasets (text-derived) \\[2pt]

TPB &
\(T(x_i)\) &
9 &
\textit{Attitude-aligned textual tone:}
\(t_1\) mean sentiment polarity;
\(t_2\) lexical valence;
\(t_3\) modality ratio. These are proxies for evaluative tone and commitment towards the post's topic or claim, not direct measures of attitude towards sharing misinformation.

\textit{Subjective norms:}
\(t_4\) group-pronoun ratio;
\(t_5\) retweet/mention token count;
\(t_6\) social-comparison cue count.

\textit{PBC:}
\(t_7\) certainty-cue ratio;
\(t_8\) instructional-language count;
\(t_9\) hashtag-directive count. &
All datasets (text-derived, see note) \\

\bottomrule

\end{tabularx}

\vspace{3pt}

\begin{minipage}{0.98\textwidth}
\footnotesize
\justifying
\noindent
\textit{Note.}
For fusion-model training, all engineered rhetorical, stance, ELM, and TPB features, including TTR, were min--max scaled using parameters estimated from the training partition. Transformer embeddings were used in their native representation. In parallel, the raw text-derived quantities required by Eq.~\eqref{eq:cps-raw} were retained for computation of the post-hoc Cognitive Propagation Score (CPS). Thus, feature scaling used for neural fusion does not alter the raw CPS values reported in Table~\ref{tab:cps_logmax}. TTR was not adjusted using a fixed-window lexical-diversity measure and is therefore interpreted with caution because it remains sensitive to post length.
\end{minipage}

\end{strip}

Figure~\ref{fig:fusion_architecture} depicts the full multi-branch fusion architecture, highlighting how each modality contributes distinct informational cues while CPS is
computed separately, post hoc, from its specified raw text-derived
quantities.

\noindent\textit{Post-hoc CPS note.}
 Figure~\ref{fig:fusion_architecture} distinguishes the learned fusion-model outputs from the Cognitive Propagation Score (CPS). CPS is not produced from the shared hidden representation \(h_i\) and is not a trainable output head. Instead, it is computed post hoc from the raw text-derived quantities specified in Eq.~\eqref{eq:cps-raw}, independently of the shared hidden representation. CPS is therefore independent of the propagation-regression target and does not use observed engagement in its construction.

For completeness, stance is represented as a 4-way encoding over support, denial, query, and comment, consistent with $S(x_i)\in\mathbb{R}^4$ in Section~\ref{sec:fusion-hidden}.

The fused vector is then passed through a dense transformation layer:
\begin{equation}
h_i = \sigma(WF(x_i) + b)
\end{equation}
where \(h_i \in \mathbb{R}^{d_h}\) denotes the shared latent representation. This representation supports two learned output heads: (1) a binary classification head for misinformation detection and (2) a regression head for propagation estimation. CPS is not generated from \(h_i\); it is calculated separately as a post-hoc index from selected raw psychology-aligned cues retained prior to feature scaling, as specified in Eq.~\eqref{eq:cps-raw}.

\subsection*{Learned Output Heads and Post-hoc CPS}

The proposed model adopts a two-task learning architecture with two learned objectives:
\begin{enumerate}
    \item Misinformation detection (binary classification),
    \item Propagation estimation (regression).
\end{enumerate}

After multi-branch feature fusion, the shared representation \(h_i \in \mathbb{R}^{d_h}\) is passed to these two learned output heads. The Cognitive Propagation Score (CPS) is subsequently computed post hoc from the raw text-derived quantities specified in Eq.~\eqref{eq:cps-raw} and does not participate in optimisation of the shared representation.

\subsubsection{Binary Classification Output}
This head predicts whether a tweet contains misinformation. It outputs a probability
\(\hat{y}^{(i)}_{\text{cls}} \in (0,1)\) via a sigmoid activation applied to a linear layer.
The classification objective is binary cross-entropy, as defined in Eq.~(24).

\begin{figure*}[t]
\centering
\refstepcounter{figure}
\label{fig:fusion_architecture}
\includegraphics[width=0.95\textwidth]{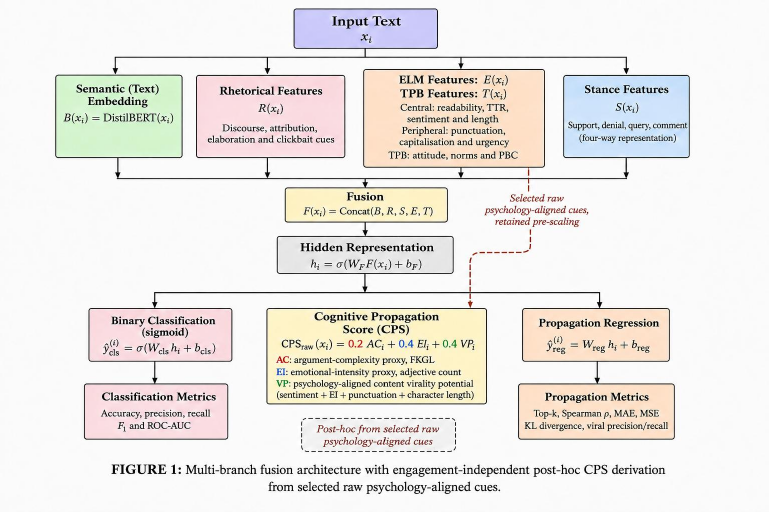}
\end{figure*}

\subsubsection{Propagation Regression Output}
This head estimates a continuous propagation target \(\hat{y}^{(i)}_{\text{reg}} \in \mathbb{R}\) using a linear activation. In other words, this head predicts an approximate engagement-intensity target as defined for each dataset (observed engagement where available, and proxy propensity otherwise).
In this paper, propagation estimation is supervised through a dedicated regression head and is methodologically distinct from CPS. For the Monkeypox dataset, \(y_{\text{eng}}^{(i)}\) represents an observed engagement-derived target. For Constraint and COVID--19\_FNIR, where
equivalent engagement fields are unavailable, the regression target is the propensity-to-spread proxy \(p_i\) defined in Section~\ref{sec:propensity_proxy}. By contrast, CPS is not a regression target and is not predicted from the shared hidden representation. It is a separate, engagement-independent post-hoc index constructed from selected
raw psychology-aligned cues, as defined in Eq.~\eqref{eq:cps-raw}. The propagation-regression objective is mean squared error, as defined in Eq.~(25).

\subsubsection{Post-hoc Cognitive Propagation Score (CPS)}
\label{subsec:cps-posthoc}

The Cognitive Propagation Score (CPS) is an engagement-independent, post-hoc index designed to summarise psychologically and rhetorically relevant cues associated with diffusion propensity. CPS is not a third neural output head and is not predicted from the shared hidden
representation \(h_i\). Instead, it is computed directly from selected raw psychology-aligned cues retained before the min--max scaling used for neural feature fusion.

To preserve the original raw-score interpretation, the post-hoc CPS
follows the scoring rule used in the original implementation. The score
combines three text-derived components: argument complexity, emotional
intensity, and content-derived virality potential. For post \(x_i\),
argument complexity is represented by the Flesch--Kincaid Grade Level,
emotional intensity is represented by the number of adjective tokens
identified through part-of-speech tagging, and virality potential is
constructed from sentiment, emotional intensity, punctuation frequency,
and raw text length.

\begin{equation}
\begin{aligned}
AC_i
&= c_1(x_i),\\
EI_i
&= \operatorname{AdjCount}(x_i),\\
VP_i
&= \frac{1}{4}\Big[
\operatorname{Sent}(x_i)
+ EI_i\\
&\qquad
+ \operatorname{PunctCount}(x_i)
+ \operatorname{CharLen}(x_i)
\Big],\\
\mathrm{CPS}_{\mathrm{raw}}(x_i)
&= 0.2\,AC_i
+ 0.4\,EI_i\\
&\qquad
+ 0.4\,VP_i .
\end{aligned}
\tag{23}
\label{eq:cps-raw}
\end{equation}

Here, \(AC_i\) denotes argument complexity and corresponds to the
Flesch--Kincaid Grade Level \(c_1(x_i)\) defined in Eq.~(1).
\(EI_i\) denotes the emotional-intensity proxy used in the original
implementation and is operationalised as the number of adjective tokens
identified through part-of-speech tagging.
\(\operatorname{Sent}(x_i)\) is the VADER compound sentiment score,
\(\operatorname{PunctCount}(x_i)\) is the number of punctuation
characters from the set \(\{.,!?\}\), and
\(\operatorname{CharLen}(x_i)\) is the raw character length of the post.
The resulting \(VP_i\) is therefore a content-derived virality-potential
component rather than a measure of observed engagement or realised
propagation.

The coefficients \(0.2\), \(0.4\), and \(0.4\) are fixed post-hoc
scoring weights assigned respectively to argument complexity, the
emotional-intensity proxy, and content-derived virality potential.
These coefficients are not learned jointly with the multi-branch model.
CPS is calculated post hoc and does not contribute an optimisation loss.

Observed engagement variables, including retweets, likes, replies, or
views, do not enter Eq.~\eqref{eq:cps-raw}. The CPS is therefore
engagement-independent by construction. Although the original
implementation used the variable name \texttt{propagation\_score} for
\(VP_i\), that quantity was calculated entirely from text-derived
sentiment, emotional-intensity, punctuation, and text-length features.
The term \emph{content-derived virality potential} is used here to avoid
confusing this component with observed propagation or with the separately
learned propensity-to-spread proxy described in Section~III-G.

Because \(\mathrm{CPS}_{\mathrm{raw}}\) is calculated from unscaled raw
feature values, it is not constrained to the interval \([0,1]\).
Consequently, its magnitude is dataset-dependent, and extreme positive
values or occasional small negative values are permissible. The raw
scale is retained because it preserves the tail behaviour reported in
Table~\ref{tab:cps_logmax}.

\paragraph{Interpretation of CPS.}
CPS should be interpreted as a theory-aligned, content-derived ranking
signal rather than as an estimate of realised platform engagement. Higher values indicate stronger weighted combinations of argument complexity, adjective-based emotional intensity, sentiment, punctuation intensity, and text length under the
CPS scoring rule. These cues are interpreted as psychology-aligned indicators of relative diffusion propensity rather than direct measurements of realised platform engagement. CPS does not estimate retweet, like, reply, or view counts.

\paragraph{Engagement independence.}
Observed engagement does not enter Eq.~\eqref{eq:cps-raw} and is not used
to fit the CPS coefficients. Where engagement metadata are available,
as in the Monkeypox dataset, associations between CPS and engagement may
be examined descriptively as an external comparison. Such engagement
information does not form part of CPS construction. The propensity-to-spread proxy \(p_i\), although engagement-supervised in the Monkeypox source domain, is likewise excluded from the CPS scoring function, thereby maintaining a strict separation between the engagement-independent CPS and the engagement-informed propagation target.

\paragraph{Role of CPS.}
CPS serves as an auxiliary behavioural signal, a propagation-oriented
content indicator, and a ranking mechanism for prioritisation and audit
when platform engagement information is incomplete, delayed, or absent.

\subsection{Propensity-to-Spread Proxy Construction}
\label{sec:propensity_proxy}

For datasets without native engagement metadata, namely Constraint and
COVID--19\_FNIR, a propensity-to-spread proxy was constructed by transferring
an engagement-supervised logistic regression model trained on the
\textit{Monkeypox dataset}. Monkeypox was used as the source domain because
it contains observed interaction variables, whereas equivalent engagement
fields are unavailable in Constraint and COVID--19\_FNIR.

\paragraph{Source-domain engagement target.}
For each Monkeypox post \(x_i\), observed engagement was defined from the
available retweet, like, and reply counts as

\[
E_i =
\log\!\left(
1 + r_i + l_i + q_i
\right),
\]

where \(r_i\), \(l_i\), and \(q_i\) denote retweets, likes, and replies,
respectively. The logarithmic transformation reduces the influence of the
long upper tail of the engagement distribution while preserving the
relative ordering of posts. Quote counts were not included in this
propensity formulation.

A binary high-engagement target was subsequently defined using the upper
10\% of the Monkeypox engagement distribution:

\[
y^{\mathrm{vir}}_i =
\mathbb{I}
\left[
E_i \geq Q_{0.90}(E)
\right],
\]

where \(Q_{0.90}(E)\) denotes the 90th-percentile engagement threshold and
\(\mathbb{I}[\cdot]\) is the indicator function. Thus,
\(y^{\mathrm{vir}}_i=1\) identifies posts in the high-engagement regime and
\(y^{\mathrm{vir}}_i=0\) identifies the remaining posts. To avoid target
leakage during model development, the percentile threshold was estimated
from the Monkeypox training partition and then applied unchanged to the
held-out partition.

\paragraph{Cross-dataset feature representation.}
The logistic model was restricted to text-derived features that could be
computed consistently across Monkeypox, Constraint, and COVID--19\_FNIR.
For each post, the common feature vector was

\[
\mathbf{z}_i =
[
z_{i1},z_{i2},\ldots,z_{i9}
],
\]

comprising post length, sentiment, affect intensity, punctuation cues,
capitalisation ratio, URL or evidence cues, hedging and modality cues,
stance cues, and rhetorical cues. The same feature definitions, extraction
rules, lexicons, and column ordering were retained across all three
datasets.

Before logistic regression fitting, the continuous proxy features were
standardised using parameters estimated from the Monkeypox training
partition only. For feature \(k\),

\[
\widetilde{z}_{ik}
=
\frac{
z_{ik}-\mu^{\mathrm{train}}_k
}{
\sigma^{\mathrm{train}}_k
},
\]

where \(\mu^{\mathrm{train}}_k\) and
\(\sigma^{\mathrm{train}}_k\) are the mean and standard deviation estimated
from the Monkeypox training data. These parameters were subsequently kept
fixed when processing the held-out Monkeypox data and the two target
datasets.

\paragraph{Logistic propensity model.}
Logistic regression was trained on Monkeypox using the standardised
cross-dataset feature vector as input and \(y^{\mathrm{vir}}_i\) as the
binary supervision target. The linear decision score is

\[
\eta_i =
\mathbf{w}^{\top}\widetilde{\mathbf{z}}_i + b,
\]

and the corresponding propensity score is

\[
p_i =
\sigma(\eta_i)
=
\frac{1}{
1+\exp(-\eta_i)
},
\qquad
p_i \in (0,1),
\]

where \(\mathbf{w}\) denotes the coefficient vector learned from Monkeypox
and \(b\) is the fitted intercept. Higher values of \(p_i\) therefore
indicate greater similarity to the text-derived profile associated with
high-engagement Monkeypox posts.

\paragraph{Transfer to engagement-missing datasets.}
After source-domain fitting and validation, the complete proxy pipeline was
frozen. Specifically, the feature definitions, feature ordering,
Monkeypox-derived standardisation parameters, logistic coefficients
\(\mathbf{w}\), and intercept \(b\) were retained without refitting.
The same nine features were then extracted from each Constraint and
COVID--19\_FNIR post, transformed using the frozen Monkeypox scaler, and
passed through the fitted logistic model:

\[
\begin{aligned}
p_i^{(D)}
&=
\sigma\!\left(
\mathbf{w}^{\top}
\widetilde{\mathbf{z}}_i^{(D)}
+b
\right),\\
D &\in
\{\text{Constraint},\,\text{COVID--19\_FNIR}\}.
\end{aligned}
\]

No engagement labels from the target datasets were used to fit or
recalibrate this model. The resulting \(p_i^{(D)}\) values therefore
represent transferred propensity-to-spread scores rather than observed
engagement.

For the propagation regression experiments reported in this paper,
\(p_i^{(D)}\) provides the continuous supervision target for Constraint and
COVID--19\_FNIR. For the Monkeypox dataset, the corresponding propagation
target is derived directly from observed engagement. Where only
prioritisation or ranking is required, the transferred scores may also be
converted to within-dataset ranks or percentiles. Because the sigmoid
function is monotonic, ranking by the linear score \(\eta_i\) or by
\(p_i\) produces the same ordering.

\paragraph{Interpretation and transfer boundary.}
The transferred score should be interpreted as an ordinal,
content-conditioned estimate of relative diffusion propensity. It does not
recover missing retweet, like, or reply counts, and it is not assumed to be
calibrated as an absolute probability of real-world virality in Constraint
or COVID--19\_FNIR. The transfer assumes only that some text-derived cues
associated with higher engagement in Monkeypox remain informative across
related health-misinformation domains. Consequently, comparisons are made
primarily within datasets rather than by treating absolute propensity
values as directly comparable across datasets.

The propensity-to-spread proxy is also distinct from the Cognitive
Propagation Score (CPS). The propensity proxy is learned under observed
Monkeypox engagement supervision and transferred to engagement-missing
datasets, whereas CPS is engagement-independent and is constructed from the
selected psychology-aligned, text-derived cues defined in Eq.~\eqref{eq:cps-raw}. The two quantities therefore provide complementary rather than interchangeable diffusion-oriented signals.

\subsection{Loss Functions}
\label{sec:loss-functions}

The model is trained in a two-task learning setting with two learned
objectives: misinformation classification and propagation regression.
CPS is calculated post hoc and therefore does not contribute a training
loss.

\paragraph{(1) Classification Loss (Binary Cross-Entropy):}
\begin{equation}
L_{\text{BCE}}
=
-\sum_{i=1}^{N}
\left[
y^{(i)}\log\hat{y}_{\text{cls}}^{(i)}
+
(1-y^{(i)})\log(1-\hat{y}_{\text{cls}}^{(i)})
\right]
\tag{24}
\end{equation}

\paragraph{(2) Propagation Regression Loss (Mean Squared Error):}
\begin{equation}
L_{\text{reg}}
=
\frac{1}{N}
\sum_{i=1}^{N}
\left(
y_{\text{eng}}^{(i)}
-
\hat{y}_{\text{reg}}^{(i)}
\right)^2
\tag{25}
\end{equation}

where \(y_{\text{eng}}^{(i)}\) denotes the dataset-specific propagation
supervision target. For the Monkeypox dataset, this target is derived from
observed engagement intensity \(E_i\). For Constraint and
COVID--19\_FNIR, where native engagement fields are unavailable, the
target is the transferred propensity-to-spread proxy \(p_i\) defined in
Section~\ref{sec:propensity_proxy}. The regression output should therefore
be interpreted relative to the dataset-specific supervision target rather
than as a universal estimate of observed platform engagement.

\paragraph{Combined Loss:}
\begin{equation}
\label{eq:loss_total}
L_{\text{total}}
=
\lambda_{\text{cls}}L_{\text{BCE}}
+
\lambda_{\text{reg}}L_{\text{reg}}
\tag{26}
\end{equation}

where
\(\lambda_{\text{cls}},\lambda_{\text{reg}}\in\mathbb{R}^{+}\)
balance the two learned objectives.

\subsubsection*{Loss Weight Selection Strategy}
The loss weights \(\lambda_{\text{cls}}\) and
\(\lambda_{\text{reg}}\) are treated as fixed hyperparameters selected
using the validation partition. Where engagement metadata are unavailable
for Constraint and COVID--19\_FNIR, \(y_{\text{eng}}\) is replaced by the
propensity-to-spread proxy \(p_i\) described in
Section~\ref{sec:propensity_proxy}. CPS does not enter \(L_{\mathrm{total}}\), because it is computed post hoc from the selected raw psychology-aligned cues defined
in Eq.~\eqref{eq:cps-raw}.

\subsection{Training Configuration \& Experimental Protocol}
Having defined the datasets, preprocessing pipeline, and feature representations used throughout this study, the following section describes the model training configuration and experimental protocol adopted to ensure robust, fair, and reproducible comparison across datasets and model families.

Table~\ref{tab:training_config} presents the model training hyperparameters. AdamW optimiser is used, together with a linear learning rate scheduler and a warm-up period, because it is a standard choice for transformer fine-tuning and provides stable optimisation with decoupled weight decay in this setting. The transformer backbone was \texttt{DistilBERT-base-uncased}, with the encoder frozen for the first five epochs and unfrozen thereafter to stabilise optimisation before full fine-tuning. Early stopping (patience = 3) was employed to prevent overfitting. Training was conducted on Google Colab using an NVIDIA A100 GPU and PyTorch v2.1.0. All experiments used a fixed random seed (seed = 42) for data splitting and model initialisation to support reproducibility.

Tweets were pre-processed, tokenised, and truncated to 128 tokens.
Engineered feature vectors used for fusion (rhetoric, stance, ELM, and
TPB) were min--max scaled, while transformer embeddings were used in
their native representation. The raw psychology-aligned quantities required by
Eq.~\eqref{eq:cps-raw} were retained separately for post-hoc CPS computation. Stratified sampling was used to split the data into 60/20/20 train--validation--test subsets. The model was trained in a two-task setting, jointly optimising misinformation classification and propagation regression. CPS was calculated separately after feature extraction and did not contribute to model optimisation.

\begin{table*}[!t]
\centering
\caption{Model Training Configuration \& Experimental Protocol}
\label{tab:training_config}
\small
\setlength{\tabcolsep}{6pt}
\renewcommand{\arraystretch}{1.10}

\begin{tabularx}{\textwidth}{@{} l X @{}}
\toprule
\textbf{Hyperparameter} & \textbf{Value} \\
\midrule
Optimiser & AdamW \\
Learning rate & $2 \times 10^{-5}$ \\
Scheduler & Linear warm-up \\
Batch size & 32 \\
Epochs & 10 \\
Early stopping & Patience = 3 (validation $L_{\text{total}}$) \\
Dropout & 0.3 \\
Max sequence length & 128 tokens \\
Loss functions &
Classification: Binary cross-entropy $L_{\text{BCE}}$ (Eq.~24);
Propagation: MSE $L_{\text{reg}}$ (Eq.~25);
Combined: $L_{\text{total}}$ (Eq.~26);
CPS: post-hoc, no optimisation loss \\
Fusion input & 768 (text) + 4 (rhetoric) + 4 (stance) + 10 (ELM) + 9 (TPB) = 795 \\
Random seed & 42 (fixed for splits and model initialisation) \\
Framework & PyTorch 2.1.0; HuggingFace Transformers \\
\bottomrule
\end{tabularx}
\end{table*}

\begin{table*}[!t]
\centering
\caption{Experimental Protocol Summary for Multi-Branch Fusion.}
\label{tab:protocol-summary-ch5}
\small
\setlength{\tabcolsep}{6pt}
\renewcommand{\arraystretch}{1.10}

\begin{tabularx}{\textwidth}{@{} l X @{}}
\toprule
\textbf{Protocol item} & \textbf{Specification} \\
\midrule
Data partitioning & Stratified 60/20/20 train--validation--test split (fixed across implemented experiments) \\
Randomisation control & Fixed seed = 42 for splits and model initialisation \\
Checkpoint selection & Best validation $L_{\text{total}}$ with early stopping (patience = 3) \\
Hyperparameter tuning & Grid search on validation only (no test-set tuning) \\
Metrics reporting & Final metrics computed on held-out test split using the selected checkpoint \\
Reported baselines & Used as contextual reference only when protocols cannot be matched \\
\bottomrule
\end{tabularx}
\end{table*}

\begin{samepage}
\Needspace{10\baselineskip}
\noindent\textbf{Training of the multi-branch fusion model proceeds as follows:}
\begin{enumerate}[label=\arabic*., leftmargin=*, nosep]
  \item Initialise all trainable model parameters \(\theta\).
  \item Fix the random seed (seed = 42) to control mini-batch shuffling
        and model initialisation.
  \item For each mini-batch, compute branch-specific representations
        \(h^{\text{text}}\), \(h^{\text{rhet}}\),
        \(h^{\text{stance}}\), and \(h^{\text{psych}}\).
  \item Concatenate the branch representations and pass the resulting
        fused representation through the shared fusion layer to obtain
        \(\hat{y}_{\text{cls}}\) and \(\hat{y}_{\text{reg}}\).
  \item Compute \(L_{\text{total}}\) using
        Equation~\eqref{eq:loss_total}.
  \item Update the trainable parameters using AdamW,
        \(\theta \leftarrow
        \theta-\eta\nabla_{\theta}L_{\text{total}}\).
  \item Repeat Steps 3--6 until convergence and retain the
        best-performing model on the validation partition.
  \item After feature extraction, compute
        \(\mathrm{CPS}_{\mathrm{raw}}\) separately using
        Equation~\eqref{eq:cps-raw}. This post-hoc computation does not
        update \(\theta\).
\end{enumerate}
\end{samepage}

\subsection{Inference Workflow}
Given a new input tweet \(x\), the model computes the following feature branches:
\begin{itemize}
    \item \(B(x)\in\mathbb{R}^{768}\): semantic (text) embedding from DistilBERT,
    \item \(R(x)\in\mathbb{R}^{4}\): rhetorical cues,
    \item \(S(x)\in\mathbb{R}^{4}\): stance representation,
    \item \(E(x)\in\mathbb{R}^{10}\): ELM psychological features,
    \item \(T(x)\in\mathbb{R}^{9}\): TPB psychological features.
\end{itemize}

These representations are concatenated and transformed according to
Eqs.~(21)--(22):

\[
\begin{aligned}
z_i
&=
\operatorname{Concat}
\left(
B(x_i),\,R(x_i),\,S(x_i),\,E(x_i),\,T(x_i)
\right),\\
h_i
&=
\phi\!\left(
W^{(1)}z_i+b^{(1)}
\right).
\end{aligned}
\]

The shared representation \(H\) is passed to two learned output heads:
\[
\hat{y}_{\text{cls}},\ \hat{y}_{\text{reg}}
\leftarrow
\text{Learned Output Heads}.
\]

Separately, the retained raw psychology-aligned cues specified
in Eq.~\eqref{eq:cps-raw} are used to calculate the post-hoc
score
\[
\mathrm{CPS}_{\mathrm{raw}}(x_i)
\leftarrow
\text{Post-hoc CPS computation}
\]
according to Eq.~\eqref{eq:cps-raw}. No hidden-state representation or
observed engagement variable is required for CPS calculation.

\subsection{Baseline Fairness \& Evaluation Protocol}
\label{subsec:baseline_fairness}

This section clarifies how evaluation comparability is handled across datasets and model families. It distinguishes the unified protocol used in the experiments in this study from the literature-reported baseline results used for the external context.

\paragraph{Evaluation protocol for implemented experiments.}
All downstream multi-branch fusion experiments in this study follow a unified protocol to minimise variance caused by data partitioning and optimisation. The source-domain logistic propensity model is fitted separately on Monkeypox as described in Section~\ref{sec:propensity_proxy}. A stratified 60/20/20 train--validation--test split was adopted to provide sufficient data for model fitting while retaining a dedicated validation partition for hyperparameter selection and a held-out test partition for final evaluation. A fixed random seed (seed = 42) was used for splitting and model initialisation to support reproducibility across runs. Model selection was based on the lowest validation
\(L_{\text{total}}\), because the architecture is trained in a two-task
setting and \(L_{\text{total}}\) provides a single criterion reflecting
joint performance across misinformation classification and propagation
regression. CPS is calculated post hoc and therefore does not influence
checkpoint selection. Early stopping with patience = 3 was used to reduce overfitting while allowing limited tolerance for short-term fluctuations in validation performance. Hyperparameters were selected using the training and validation partitions only, and the held-out test split was used once for final reporting.

\paragraph{Literature baselines and comparability boundary.}
Baseline results for Constraint, COVID--19\_FNIR, and Monkeypox datasets are drawn from peer-reviewed literature and are included to contextualise performance on each benchmark. Because published studies can differ in split protocols, preprocessing, feature access, and tuning regimes, literature results are treated as contextual reference rather than as strictly protocol-matched baselines. Where protocols cannot be aligned, strong cross-paper claims are avoided based on small metric differences and such comparisons should be interpreted cautiously.

\paragraph{Target definitions for propagation-oriented evaluation.}
Propagation-oriented supervision depends on engagement availability. When engagement traces are available (e.g., Monkeypox dataset), propagation targets are defined from observed engagement. When engagement fields are unavailable (Constraint and COVID--19\_FNIR), propagation supervision uses the propensity-to-spread proxy defined in Section~\ref{sec:propensity_proxy}. This maintains a consistent evaluation objective within each dataset, while avoiding claims that proxy propensity is engagement-equivalent virality.

\paragraph{Fairness of information access.}
To avoid privileged-information advantages in detection comparisons, the proposed model uses only the declared text-derived branches (semantic, rhetorical, stance, ELM, TPB). Observed engagement is not used as an input to the misinformation-detection
model or in the post-hoc CPS calculation. Where available, engagement is
used only for propagation-regression supervision, auxiliary comparison,
and proxy validation as specified for the corresponding experiment.

\paragraph{Mpox benchmark comparability.}
Mpox results are sensitive to the unit of analysis. Claim-level benchmarks such as PoxVerifi~\cite{kolluri2022poxverifi} are not directly comparable to tweet-level classification, so such results are discussed as contextual evidence rather than as directly comparable baselines in the main tables.

\subsection{Evaluation Metrics}
\label{sec:eval_metrics}

The detection and diffusion-oriented components are evaluated using
classification, regression, ranking, and distributional metrics.
For misinformation detection, Accuracy, Precision, Recall,
\(F_1\)-score, and ROC--AUC are reported. For propagation prediction,
Mean Squared Error (MSE) and Mean Absolute Error (MAE) quantify
prediction error, while Spearman's rank correlation coefficient
(\(\rho\)) assesses agreement between predicted and target propagation
rankings. KL divergence quantifies divergence between predicted and
target propagation distributions.

CPS is not evaluated as a prediction task because it is calculated
deterministically post hoc. Instead, the raw CPS distribution is
characterised using the mean, standard deviation, minimum, maximum, and
\(\log_{10}\) of the maximum where appropriate to aid interpretation of
highly skewed upper tails. Unless otherwise stated, lower MSE, MAE, and
KL divergence indicate better propagation prediction, whereas higher
Accuracy, Precision, Recall, \(F_1\), ROC--AUC, and Spearman's
\(\rho\) indicate better predictive or ranking performance.

\section{Results}
This section presents the study's findings across two learned tasks, misinformation classification and propagation prediction, together with a separate post-hoc analysis of the Cognitive Propagation Score (CPS).

\noindent\textit{Boundary.} Where baseline values are drawn from prior studies, they are cited for context and interpreted cautiously because evaluation protocols are not fully standardised across papers.

\subsection{Classification Performance}
Table~\ref{tab:classification} reports accuracy, precision, recall, F1-score and ROC-AUC. COVID--19\_FNIR achieves very high classification scores (Table~\ref{tab:classification}), which is consistent with repeated rumour templates and relatively stable linguistic patterns reported for COVID-related misinformation \cite{aral2020hype}. However, these ceiling-level metrics should be interpreted as within-dataset performance under the current evaluation protocol, rather than evidence of guaranteed out-of-domain generalisation. The Constraint dataset also performs strongly, indicating distinctive textual cues \cite{shu2020socialcontext}. Monkeypox dataset yields lower performance (approximately 80\% accuracy), plausibly reflecting partial truths and emotive framing that blur class boundaries \cite{chou2020misinformation}.

\begin{table}[t]
\centering
\caption{Classification performance across the three datasets.}
\label{tab:classification}

\small
\setlength{\tabcolsep}{4pt}
\renewcommand{\arraystretch}{1.08}

\begin{tabular}{@{}lccc@{}}
\toprule
\textbf{Metric}
&
\shortstack{\textbf{Constraint}\\\textbf{dataset}}
&
\shortstack{\textbf{Monkeypox}\\\textbf{dataset}}
&
\shortstack{\textbf{COVID--19}\\\textbf{FNIR}}
\\
\midrule

Accuracy
& 0.9781
& 0.8042
& \textbf{0.9929}
\\

Precision
& 0.9600
& 0.7851
& \textbf{0.9892}
\\

Recall
& 0.9954
& 0.8326
& \textbf{0.9973}
\\

F1
& 0.9774
& 0.8082
& \textbf{0.9932}
\\

ROC--AUC
& 0.9805
& 0.8858
& \textbf{0.9999}
\\

\bottomrule
\end{tabular}
\end{table}

\subsection{Ablation Study: Contribution of Fusion Branches}
\label{sec:ablation}

To evaluate the relative impact of each modality in the fusion architecture, an ablation study was conducted by selectively disabling individual branches and retraining the model using identical hyperparameters and the same fixed train--validation--test split. Classification performance was recalculated for the Constraint, Monkeypox, and COVID--19\_FNIR datasets. Table~\ref{tab:ablation} reports F1-score and ROC-AUC, alongside the relative drop compared to the full fusion configuration.

\paragraph{Ablation analysis.}
To quantify the marginal contribution of each feature branch, the full fusion model is compared with variants where one branch is removed. Here, the psychological branch is defined as the concatenation of ELM and TPB features \((10 + 9)\):
\begin{enumerate}[label=\alph*., leftmargin=*]
  \item for each branch \(b \in \{\text{text}, \text{rhetoric}, \text{stance}, \text{psych}\}\), train a model with branch \(b\) omitted under the same split and hyperparameters as the full model, where \(\text{psych}=\text{Concat}(\text{ELM},\text{TPB})\),
  \item compute \(\Delta F1_b = F1_{\text{full}} - F1_{-b}\) (and analogously for ROC-AUC),
  \item summarise the resulting performance drops in Table~\ref{tab:ablation}.
\end{enumerate}

\begin{table*}[t]
\centering
\caption{Ablation study results by dataset, reporting F1-score and ROC--AUC for each ablated model. Relative percentage drops from the full fusion model are shown in parentheses for each metric.}
\label{tab:ablation}
\setlength{\tabcolsep}{4pt}
\renewcommand{\arraystretch}{1.15}
\small
\begin{tabularx}{\textwidth}{
  >{\raggedright\arraybackslash}p{0.26\textwidth}
  >{\centering\arraybackslash}X
  >{\centering\arraybackslash}X
  >{\centering\arraybackslash}X
  >{\centering\arraybackslash}X
  >{\centering\arraybackslash}X
  >{\centering\arraybackslash}X}
\toprule
& \multicolumn{2}{c}{\textbf{Constraint}} & \multicolumn{2}{c}{\textbf{Monkeypox}} & \multicolumn{2}{c}{\textbf{COVID--19\_FNIR}} \\
\cmidrule(lr){2-3} \cmidrule(lr){4-5} \cmidrule(lr){6-7}
\textbf{Feature Removed} & \textbf{F1-score} & \textbf{ROC--AUC} & \textbf{F1-score} & \textbf{ROC--AUC} & \textbf{F1-score} & \textbf{ROC--AUC} \\
\midrule
None (Full Fusion)
& 0.9774 & 0.9805
& 0.8082 & 0.8858
& 0.9932 & 0.9999 \\

Rhetorical Branch
& 0.9141 (\(\downarrow\)6.5\%) & 0.9752 (\(\downarrow\)0.5\%)
& 0.7986 (\(\downarrow\)1.2\%) & 0.8733 (\(\downarrow\)1.4\%)
& 0.9901 (\(\downarrow\)0.3\%) & 0.9976 (\(\downarrow\)0.2\%) \\

Stance Branch
& 0.9068 (\(\downarrow\)7.2\%) & 0.9634 (\(\downarrow\)1.7\%)
& 0.7819 (\(\downarrow\)3.2\%) & 0.8561 (\(\downarrow\)3.4\%)
& 0.9872 (\(\downarrow\)0.6\%) & 0.9961 (\(\downarrow\)0.4\%) \\

Psychological Branch (ELM + TPB)
& 0.8927 (\(\downarrow\)8.7\%) & 0.9510 (\(\downarrow\)3.0\%)
& 0.7665 (\(\downarrow\)5.2\%) & 0.8442 (\(\downarrow\)4.7\%)
& 0.9783 (\(\downarrow\)1.5\%) & 0.9944 (\(\downarrow\)0.6\%) \\

Semantic (Text) Branch (DistilBERT)
& 0.8342 (\(\downarrow\)14.7\%) & 0.9096 (\(\downarrow\)7.2\%)
& 0.7201 (\(\downarrow\)10.9\%) & 0.8013 (\(\downarrow\)9.5\%)
& 0.9026 (\(\downarrow\)9.1\%) & 0.9672 (\(\downarrow\)3.3\%) \\
\bottomrule
\end{tabularx}

\vspace{2pt}
\begin{minipage}{0.98\textwidth}
\footnotesize
\textbf{Note:} Values in parentheses denote the relative percentage drop of each metric compared with the full fusion model for the same dataset.
\end{minipage}
\end{table*}

Overall, the semantic (text) branch, comprising the DistilBERT representation \(B(x_i)\), is the primary driver of performance across datasets. Its removal produces the largest declines in
\(F_1\)-score and ROC--AUC, particularly for the Monkeypox and COVID--19\_FNIR datasets. The psychological branch (ELM + TPB) yields consistent degradation when removed, most notably on the Monkeypox dataset and Constraint dataset, indicating that behavioural and affective cues improve discrimination for ambiguous or emotionally framed content. Rhetorical and stance branches typically produce smaller drops on Monkeypox and COVID--19\_FNIR datasets, but their removal is more consequential on the Constraint dataset, suggesting that style and stance cues help separate classes in that corpus.

\subsection{Propagation Prediction}
\label{sec:propagation_prediction}
All propagation results in this subsection are conditional on the evaluation set-up, including the train--test split, the virality definition and thresholding rule, and the Top-$k$ retrieval protocol. Accordingly, Top-5 and viral precision and recall should be interpreted as performance under this specific set-up rather than as a general guarantee.

Table~\ref{tab:propagation} summarises eight propagation metrics. COVID--19\_FNIR records the lowest errors, a Top-5 hit rate of 1.000, and low KL divergence (Table~\ref{tab:propagation}), indicating a relatively predictable proxy-defined diffusion-risk profile under the chosen virality definition and thresholds. Monkeypox dataset achieves very strong rank ordering and high viral precision, although recall (62.6\%) suggests some borderline posts fall below the viral threshold and are not retrieved. Constraint dataset shows good absolute error (MSE \(\sim\) 0.0019) but fails on viral recall (0) and Top-5 hits. Because the Constraint dataset lacks native engagement metadata, these propagation metrics are computed against the propensity proxy target, and the results suggest that the proxy distribution is outlier-heavy and that the viral threshold \(\tau\) is miscalibrated for extreme values. This aligns with prior work showing that diffusion-related signals in social systems can be heavy-tailed \cite{bovet2019influence}.

\paragraph{Proxy-based propagation supervision and evaluation.}
Propagation evaluation uses an engagement supervision target \(y_{\text{eng}}\). For the Monkeypox dataset, \(y_{\text{eng}}\) is derived from native engagement metadata. For Constraint and COVID--19\_FNIR datasets, where engagement fields are unavailable, \(y_{\text{eng}}\) is replaced by a propensity score proxy \(p^{(i)}\) computed as described in Section~\ref{sec:propensity_proxy}. Accordingly, all propagation metrics in Table~\ref{tab:propagation} for Constraint and COVID--19\_FNIR datasets are computed against the propensity proxy target rather than raw platform engagement.

\begin{table*}[!t]
\centering

\caption{Propagation metrics across the three datasets.}
\label{tab:propagation}

\small
\setlength{\tabcolsep}{6pt}
\renewcommand{\arraystretch}{1.08}

\begin{tabular}{
    >{\raggedright\arraybackslash}p{5.0cm}
    >{\centering\arraybackslash}p{3.2cm}
    >{\centering\arraybackslash}p{3.2cm}
    >{\centering\arraybackslash}p{3.2cm}
}
\toprule

\multicolumn{1}{l}{\textbf{Metric}}
&
\multicolumn{1}{c}{\shortstack{\textbf{Constraint}\\\textbf{dataset}}}
&
\multicolumn{1}{c}{\shortstack{\textbf{Monkeypox}\\\textbf{dataset}}}
&
\multicolumn{1}{c}{\textbf{COVID--19\_FNIR}}
\\

\midrule

MSE (propagation)
& 0.0019
& 0.0029
& \textbf{0.0003}
\\

MAE (propagation)
& 0.0250
& 0.0458
& \textbf{0.0129}
\\

Spearman's rho (\(\rho\))
& 0.2272
& 0.9952
& \textbf{0.9954}
\\

Top-5 hit rate
& 0.0000
& 0.8000
& \textbf{1.0000}
\\

Viral threshold
& \(\sim 0.0453\)
& \(\sim 1.3041\)
& \(\sim 0.5667\)
\\

Precision (viral)
& 0.0000
& 0.9840
& \textbf{0.9858}
\\

Recall (viral)
& 0.0000
& 0.6259
& \textbf{0.8742}
\\

KL divergence
& 2.5319
& 6.0333
& \textbf{0.4045}
\\

\bottomrule
\end{tabular}

\vspace{2pt}

\begin{minipage}{14.6cm}
\footnotesize
\justifying
\noindent
\textbf{Notes for Table~\ref{tab:propagation}:}
For the Constraint and COVID--19\_FNIR datasets, the propagation target is the propensity-score proxy \(p^{(i)}\) described in Section~\ref{sec:propensity_proxy}; for the Monkeypox dataset, the target is engagement-derived. A post is labelled viral when \(y_{\mathrm{eng}}^{(i)} \geq \tau\), where \(\tau\) is defined from the corresponding target distribution. Lower values indicate better performance for MSE, MAE, and KL divergence; higher values indicate better performance for Spearman's correlation, precision, recall, and the top-5 hit rate. Bold type identifies the best value in each row where direct comparison is applicable.
\end{minipage}

\end{table*}

The Constraint dataset's viral threshold lies far below a small set of extreme outliers, which helps explain the absence of Top-5 hits and the resulting near-zero viral precision and recall. The higher KL divergence ($2.53$) indicates a pronounced distributional mismatch. The Monkeypox dataset combines high rank correlation with high viral precision, but its elevated KL divergence suggests that distributional alignment remains challenging under the current virality definition. COVID--19\_FNIR combines minimal error, ceiling-level Top-5 retrieval, and well-balanced precision and recall, a pattern consistent with repeated rumour templates and comparatively stable user reactions \cite{aral2020hype}.

\subsection{Cognitive Propagation Score (CPS) Evaluation}
The raw post-hoc CPS combines argument complexity,
an adjective-based emotional-intensity proxy, and
content-derived virality potential constructed from sentiment,
emotional intensity, punctuation count, and text length, as
defined in Eq.~\eqref{eq:cps-raw}. These quantities are treated
as selected psychology-aligned cues within the broader ELM
and TPB theoretical framing of the study. Table~\ref{tab:cps_logmax}
reports the raw CPS distribution for each dataset. Because these raw
scores retain the original feature magnitudes, they are not restricted
to \([0,1]\) and may exhibit pronounced upper-tail skew. For descriptive
purposes, \(\log_{10}\) is additionally reported for the maximum raw CPS
value; the transformation is not applied to the underlying CPS scores.

\begin{table*}[!t]
\centering

\caption{\textbf{Raw post-hoc CPS statistics across datasets, with
$\log_{10}$ additionally reported for the maximum value to characterise
upper-tail skew.}}
\label{tab:cps_logmax}

\small
\setlength{\tabcolsep}{12pt}
\renewcommand{\arraystretch}{1.10}

\begin{tabular}{@{}lccc@{}}
\toprule

\textbf{Metric}
&
\textbf{Constraint dataset}
&
\textbf{Monkeypox}
&
\textbf{COVID--19\_FNIR}
\\

\midrule

CPS Mean
& \textbf{23.07}
& 18.77
& 21.18
\\

CPS Std
& \textbf{20.02}
& 10.10
& 10.77
\\

CPS Min
& 1.86
& 0.00
& \textbf{-0.10}
\\

CPS Max
& \textbf{1257.78}
& 57.02
& 63.78
\\

$\log_{10}(\text{CPS Max})^\dagger$
& \textbf{3.10}
& 1.76
& 1.81
\\

\bottomrule
\end{tabular}

\vspace{0.5ex}

\begin{minipage}{0.90\textwidth}
\footnotesize
\justifying
\noindent
\textbf{Notes for Table~\ref{tab:cps_logmax}:}
All values labelled CPS Mean, CPS Std, CPS Min, and CPS Max are reported
on the raw, unnormalised post-hoc CPS scale defined in
Eq.~\eqref{eq:cps-raw}. They are not min--max-scaled CPS values.
Accordingly, the score is not constrained to \([0,1]\), and occasional
negative values and large positive outliers are permissible.
Bold indicates the most extreme value per row, maximum for Mean, Std,
Max, and \(\log_{10}(\mathrm{Max})\), and minimum for Min.
The \(\log_{10}\) transformation is applied only to the reported raw
CPS maximum and does not alter the CPS values used in the analysis.
\end{minipage}

\end{table*}

The Constraint dataset shows a heavy-tailed raw CPS distribution
(max \(\approx 1{,}258\)), indicating that a small number of posts
contain exceptionally strong combinations of the cues represented in
the post-hoc CPS. These instances can therefore be treated as
high-priority candidates under the CPS ranking function, rather than as
posts with known or predicted realised reach
\cite{bovet2019influence}. The pronounced upper tail is consistent with
the broader observation that diffusion-related phenomena can be strongly
skewed, with a small number of items dominating observed spread
\cite{vosoughi2018spread}. Because CPS is a theory-derived post-hoc
content index rather than an observed platform variable, these values are
interpreted as relative diffusion propensity under the CPS scoring rule,
not as realised engagement.

The Monkeypox dataset exhibits a tighter CPS band (max $\approx 57$) and includes near-zero CPS scores, suggesting that many posts contain weaker combinations of the selected psychology-aligned cues used by the engagement-independent CPS definition, and therefore are assessed as low diffusion propensity under the same scoring function. COVID--19\_FNIR sits between these extremes, with moderate variance and occasional negative CPS values, which can reflect sceptical or corrective framing. Taken together, these contrasts imply tailored mitigation, outlier-aware filtering for tail-heavy regimes like the Constraint dataset, simpler cut-offs for bounded regimes like the Monkeypox dataset, and periodic recalibration for COVID-like domains. Interpreted as diffusion-risk profiles, the heavier CPS upper tail in Constraint indicates a greater concentration of rare, high-diffusion-cue outliers under the CPS scoring function, whereas the more bounded CPS range in the Monkeypox dataset indicates fewer posts exhibiting the high-CPS cue combinations involving argument complexity, emotional-intensity, sentiment, punctuation, and text-length signals captured by the CPS definition.

\subsection{Dataset-Level Analysis: Classification, Propagation, and CPS Profiles}
The integrated analysis of classification metrics (Table~\ref{tab:classification}), propagation metrics (Table~\ref{tab:propagation}), and CPS distributions (Table~\ref{tab:cps_logmax}) highlights how misinformation detection and diffusion properties vary across Constraint, COVID--19\_FNIR and the Monkeypox datasets. Consistent with Lazer et al. \cite{lazer2018science}, Shu et al. \cite{shu2020socialcontext}, and Vosoughi et al. \cite{vosoughi2018spread}, datasets can exhibit either tail-heavy diffusion with sporadic super-viral items, or more uniform patterns shaped by cognitive and social forces.

\textbf{Constraint Dataset.}
Classification is strong, with accuracy of 97.81\% and F1-score of 97.74\% (Table~\ref{tab:classification}), indicating that misleading content contains distinctive cues that the model exploits \cite{shu2020socialcontext}. Propagation metrics, however, indicate divergence, with Spearman’s correlation of 0.2272 and KL divergence of 2.5319 (Table~\ref{tab:propagation}). This suggests that strong classification performance does not necessarily translate to accurate virality retrieval under a single global threshold in an outlier-heavy regime \cite{bovet2019influence}.

\textbf{CPS for Constraint Dataset.}
As shown in Table~\ref{tab:cps_logmax}, CPS exhibits a mean of 23.07 and a maximum of 1257.78, indicating substantial tail risk. This pattern is consistent with empirical findings that diffusion can be dominated by rare extreme cascades \cite{vosoughi2018spread}.

\textbf{Monkeypox Dataset.}
Classification is lower than COVID--19\_FNIR and Constraint dataset (F1 = 80.82\%, Table~\ref{tab:classification}), consistent with partial truths and subtler framing \cite{chou2020misinformation}. Propagation prediction shows excellent rank ordering (Spearman’s \(\rho \approx 0.9952\)), high viral precision, and moderate recall, but the higher KL divergence indicates remaining distributional mismatch (Table~\ref{tab:propagation}).

\textbf{CPS for Monkeypox Dataset.}
CPS is comparatively bounded (max = 57.02) with many low or zero values (Table~\ref{tab:cps_logmax}), suggesting fewer extreme outliers and more constrained diffusion potential.

\textbf{COVID--19\_FNIR Dataset.}
Classification performance is near ceiling (F1 = 99.32\%, ROC-AUC = 0.9999, Table~\ref{tab:classification}), consistent with recurrent misinformation templates and stable textual markers \cite{aral2020hype}. For propagation-oriented evaluation, COVID--19\_FNIR does not provide native engagement traces, therefore the regression head is supervised and assessed against the propensity-to-spread proxy $p_i$ (Section~\ref{sec:propensity_proxy}). Under this proxy-defined target, the model achieves very strong ranking agreement ($\rho = 0.9954$), minimal error (MSE = 0.0003; MAE = 0.0129), a Top-5 hit rate of 1.000, and low distribution divergence (KL = 0.4045; Table~\ref{tab:propagation}). These results should be interpreted as alignment with the proxy diffusion-risk objective within COVID--19\_FNIR, rather than as prediction of observed platform engagement.

\textbf{CPS for COVID--19\_FNIR.}
CPS exhibits moderate variance with a small negative minimum (-0.10), which can reflect sceptical or corrective framing that reduces persuasive appeal (Table~\ref{tab:cps_logmax}). Overall, COVID--19\_FNIR presents relatively predictable diffusion under the current evaluation set-up.

\subsection{Integrated Interpretation: Fusion Effectiveness and Theoretical Relevance}
\paragraph{Fusion Model Consistency.}
Across all datasets, the multi-branch fusion model yields strong classification results and competitive propagation prediction under the defined evaluation protocol. Its strength lies in leveraging heterogeneous cues, textual, rhetorical, stance, and psychological, that are complementary rather than redundant. The ablation results in Table~\ref{tab:ablation} confirm that removing any branch reduces performance, with the largest drops arising when the semantic (text) branch is removed.

\paragraph{Psychological Features and Interpretability.}
The inclusion of ELM and TPB-inspired features supports a more interpretable account of why content is classified as misinformation, particularly in datasets where persuasion-style cues are salient \cite{petty1986elaboration, ajzen1991theory}. CPS is framed as a theory-driven auxiliary metric, not a ground-truth substitute, that provides an additional lens on potential diffusion risk.

\paragraph{Summary.}
Overall, the results indicate that fusion improves robustness by combining semantic and psychologically grounded signals, while dataset-specific diffusion regimes still require tailored thresholding and evaluation choices.

\paragraph{Cognitive Propagation Score (CPS) as Auxiliary Signal.}
CPS is not framed as a replacement for engagement-based virality ground truth, but as a theory-driven auxiliary metric introduced in this paper. It is intended to summarise psychologically motivated cues, so that diffusion risk can still be discussed when engagement annotations are sparse or delayed. Although CPS is not validated through human studies in the present work, its skewed, heavy-tailed profiles align with prior evidence that online diffusion is often dominated by rare, highly amplified outliers \cite{vosoughi2018spread}.

\paragraph{Constraint dataset vs. Monkeypox dataset Discrepancy.}
The weaker performance on the Monkeypox dataset, relative to COVID--19\_FNIR and the Constraint dataset, likely reflects dataset and signal characteristics rather than a single modelling limitation. Mpox posts are shorter and more homogeneous, which reduces the variance of rhetorical and psychological cues available to the fusion branches. In addition, the comparatively bounded CPS range (Table~\ref{tab:cps_logmax}) indicates fewer extreme high-scoring instances under the engagement-independent CPS definition, limiting outlier-driven separation signals. By contrast, the Constraint dataset exhibits a more tail-heavy CPS distribution under the same engagement-independent CPS definition, indicating that a small number of posts contain unusually strong combinations of ELM and TPB cues that dominate the upper tail and can disproportionately influence ranking behaviour and prioritisation outcomes.

\paragraph{Propagation Predictability \& Distributional Alignment.}
Propagation predictability differs by metric and dataset. The Monkeypox and COVID--19\_FNIR datasets both show near-ceiling rank ordering (Spearman's $\rho \approx 0.995$, Table~6), but the underlying targets differ. For Monkeypox, this reflects agreement with observed engagement intensity, whereas for COVID--19\_FNIR it reflects agreement with the transferred propensity-to-spread proxy. KL divergence remains higher for Monkeypox than for COVID--19\_FNIR, indicating greater difficulty in matching the observed engagement distribution in the former. The Constraint dataset shows weak rank correlation and zero viral recall under the current thresholding rule, consistent with an outlier-dominated proxy-target regime in which a single global threshold is misaligned with the distribution tail \cite{bovet2019influence}.

\paragraph{Implications for Misinformation Surveillance.}
Together, the results indicate that integrating psychological theory into NLP models can strengthen interpretability and can provide additional signals for reasoning about diffusion risk alongside textual features. In settings where engagement data are sparse or delayed, CPS may be useful as a supplementary cue, but it should not be treated as ground truth. Future work should validate CPS through human-centred evaluation, for example, share-intent ratings, expert review, or controlled attention studies.

\paragraph{Summary.}
Overall, the fusion model outperforms strong baselines on two of the three datasets, and shows interpretable CPS patterns aligned with persuasion theory. Performance remains dataset-dependent, with reduced effectiveness on the Monkeypox dataset, indicating that dataset structure, cue variance, and diffusion regime influence the suitability of fusion-based approaches.

\subsection{Comparison with Baselines}
\label{sec:baseline_comparison}

To contextualise the effectiveness of the proposed multi-branch fusion model, results are compared against representative \emph{reported} baselines from peer-reviewed studies on the same benchmark datasets. Because published work can differ in split protocols, preprocessing, and tuning regimes, these comparisons are interpreted as contextual rather than as strictly protocol-matched head-to-head evaluations. The numerical differences reported below therefore indicate relative performance magnitudes, while acknowledging potential protocol effects.

On the \textbf{Constraint} dataset, our model achieved an F1-score of 97.74\% and an accuracy of 97.81\%. Relative to the CNN+BiLSTM results reported by Sharif et al.~\cite{sharif2021combating} (92.01\% F1 and 92.01\% accuracy), this corresponds to differences of +5.73 F1 and +5.80 percentage points under the respective reported evaluation settings.

For the \textbf{COVID--19\_FNIR} dataset, the fusion model achieved 99.32\% F1 and 99.29\% accuracy. Compared with the baseline reported in an earlier study co-authored by the first author \cite{sikosana2024hybrid}, which achieved an F1-score of 97.60\% and accuracy of 97.69\%, the corresponding differences are $+1.72$ and $+1.60$ percentage points.

In contrast, on the \textbf{Monkeypox} dataset, the fusion model attained 80.82\% F1 and 80.42\% accuracy, compared to 90.57\% F1 and 90.41\% accuracy reported by Mohbey et al.~\cite{mohbey2024cnn}. The differences are -9.75 F1 and -9.99 percentage points, suggesting that short and syntactically sparse posts may reduce the benefit of the current fusion branches, and that sequential inductive biases can remain competitive in this setting.

Overall, Table~\ref{tab:baseline_comparison} shows that the proposed model exceeds reported baselines on Constraint and COVID--19\_FNIR, while underperforming on the Monkeypox dataset, consistent with dataset-specific differences in text length, cue variance, and topic homogeneity.

\begin{table*}[!t]
\centering

\caption{Baseline vs. Fusion Performance}
\label{tab:baseline_comparison}

\small
\setlength{\tabcolsep}{5pt}
\renewcommand{\arraystretch}{1.15}

\begin{tabularx}{\textwidth}{
    >{\raggedright\arraybackslash}p{2.2cm}
    >{\raggedright\arraybackslash}X
    >{\centering\arraybackslash}p{1.25cm}
    >{\centering\arraybackslash}p{1.25cm}
    >{\centering\arraybackslash}p{1.45cm}
    >{\centering\arraybackslash}p{1.35cm}
    >{\centering\arraybackslash}p{1.15cm}
    >{\centering\arraybackslash}p{1.15cm}
}
\toprule

\textbf{Dataset} &
\textbf{Reported baseline} &
\textbf{Accuracy} &
\textbf{F1-score} &
\textbf{Fusion Accuracy} &
\textbf{Fusion F1-score} &
\(\boldsymbol{\Delta}\) \textbf{F1-score} &
\(\boldsymbol{\Delta}\) \textbf{Accuracy} \\

\midrule

Constraint &
CNN+BiLSTM~\cite{sharif2021combating} &
92.01\% &
92.01\% &
97.81\% &
97.74\% &
\(\uparrow\) 5.73 &
\(\uparrow\) 5.80 \\

COVID--19\_FNIR &
RoBERTa+RoBERTa~\cite{sikosana2024hybrid} &
97.69\% &
97.60\% &
99.29\% &
99.32\% &
\(\uparrow\) 1.72 &
\(\uparrow\) 1.60 \\

Monkeypox &
CNN+LSTM~\cite{mohbey2024cnn} &
90.41\% &
90.57\% &
80.42\% &
80.82\% &
\(\downarrow\) 9.75 &
\(\downarrow\) 9.99 \\

\bottomrule
\end{tabularx}

\vspace{3pt}

\begin{minipage}{0.97\textwidth}
\footnotesize
\justifying
\noindent
\textit{Note.}
Baseline values are taken from the cited papers. The study by Sikosana et al.~\cite{sikosana2024hybrid} was co-authored by the first author of the present study. \(\Delta\) values represent differences between the fusion-model results and the reported baseline values and should be interpreted contextually because evaluation protocols may differ.
\end{minipage}

\end{table*}
\section{Discussion}
This study introduced a multi-branch fusion model that integrates psychologically grounded signals from the Elaboration Likelihood Model (ELM) and the Theory of Planned Behaviour (TPB) with rhetorical, stance, and contextual textual representations. The goal was to improve health misinformation detection, examine diffusion-oriented behaviour under the available supervision regime, and provide an interpretable estimate of cognitive diffusion propensity via the Cognitive Propagation Score (CPS). Overall, the results indicate that multi-perspective feature fusion can improve classification performance and can support interpretation of model behaviour across datasets with differing linguistic and diffusion-risk characteristics.

\subsection{Interpretable Gains from Multi-Branch Fusion}
The findings suggest that combining psychological, linguistic, and stance cues yields measurable and interpretable gains. Transformer-based models are effective for factuality and misinformation detection \cite{devlin2019bert, liu2019roberta}, but they typically provide limited behavioural interpretability when used alone. In contrast, the proposed approach explicitly operationalises ELM and TPB constructs \cite{ajzen1991theory, petty1986elaboration}, which remains relatively uncommon in mainstream NLP misinformation pipelines, and this design choice aims to improve both predictive performance and interpretive transparency.

The ablation results provide direct evidence of complementary contributions across branches (Table~\ref{tab:ablation}). Removing the semantic (text) branch, represented by DistilBERT,
produces the largest drop in performance, confirming that contextual semantics remain foundational for detection. However, removing the psychological branch (ELM and TPB) produces consistent, non-trivial reductions in $F_1$ across datasets, indicating that psychologically motivated cues add information beyond what is captured by the transformer embedding alone. This contribution is plausibly strongest when posts rely on affective framing, normative appeals, certainty cues, or heuristic signals that are not fully resolved by surface semantics.

Stance and rhetorical branches contribute smaller but systematic improvements, consistent with prior work showing that stance and discourse structure can strengthen rumour and misinformation classification, particularly when semantics alone are insufficient to resolve credibility \cite{zubiaga2018detection, kumar2019tree}. Collectively, these findings support the view that heterogeneous evidence streams can be complementary within fusion architectures rather than redundant.

\subsection{Dataset-Specific Propagation Patterns: Outliers, Norms, and Heuristics}
The propagation-oriented results suggest that diffusion-risk behaviour depends both on dataset characteristics and on the evaluation design, including how the propagation target is defined and how concentrated or dispersed the target distribution is. In this study, two propagation settings are distinguished: observed engagement targets where platform traces are available, and propensity-to-spread proxy targets where engagement fields are absent. Accordingly, propagation-related findings for Constraint and COVID--19\_FNIR should be interpreted as proxy-based diffusion-risk estimates rather than as direct measurements of observed platform engagement.

COVID--19\_FNIR exhibits highly predictable rank ordering under the propensity proxy, with strong Spearman correlation and low error (Table~\ref{tab:propagation}). In practical terms, this means that the model was effective at reproducing the relative ordering implied by the proxy target within this dataset, even though no native engagement traces were available. This result should therefore be interpreted as strong alignment with the proxy-defined diffusion-risk signal, rather than as direct prediction of real-world platform engagement. One plausible explanation is that COVID--19\_FNIR contains recurring misinformation templates and comparatively stable linguistic patterns, making the text-derived propagation proxy more internally consistent and therefore easier for the model to learn.

The Constraint dataset shows a different regime. Despite strong classification performance, it exhibits weaker rank predictability and higher distribution divergence, which is consistent with a tail-dominated proxy target in which a small number of extreme instances disproportionately influence distributional alignment \cite{vosoughi2018spread, bovet2019influence}. In such settings, psychologically informed cues related to norms and perceived behavioural control can provide additional explanatory value for prioritisation behaviour, even when global distribution matching remains difficult \cite{ajzen1991theory}.

The Monkeypox dataset provides a contrast because engagement traces are available and the propagation target is engagement-derived. It exhibits relatively strong rank correlation but higher KL divergence than COVID--19\_FNIR (Table~\ref{tab:propagation}), suggesting that the model can order instances reasonably well while still struggling to align the full engagement distribution. This is consistent with a more bounded engagement regime and the fact that distributional matching is more demanding than rank agreement. Any interpretation about cognitive routes remains tentative because route processing is not directly measured in this study \cite{chou2020misinformation}.

\subsection{Cognitive Propagation Score (CPS): Behaviourally Grounded Interpretability}
A central contribution of this paper is CPS, a theory-motivated
auxiliary metric situated within the broader ELM and TPB framing of
the study \cite{petty1986elaboration,ajzen1991theory}. Its post-hoc
scoring rule operationalises selected psychology-aligned cues comprising
argument complexity, adjective-based emotional intensity, sentiment,
punctuation, and text length. These cues provide interpretable
content-level signals that are theoretically consistent with cognitive,
affective, and heuristic mechanisms implicated in persuasion and
behavioural response \cite{petty1986elaboration,ajzen1991theory}.
Unlike diffusion studies centred on observed platform propagation
patterns and social amplification \cite{vosoughi2018spread,bovet2019influence},
CPS therefore provides an interpretable, engagement-independent lens
for discussing relative diffusion propensity within the scope of
psychologically motivated, text-derived features.

CPS is engagement-independent by construction and is calculated
post hoc from the selected raw psychology-aligned cues specified
in Eq.~\eqref{eq:cps-raw}. Observed engagement fields do
not enter the CPS scoring function, and CPS is not derived from the
propagation-regression target or the shared hidden representation. Where
engagement traces are available, as in the Monkeypox dataset, associations
between \(\mathrm{CPS}_{\mathrm{raw}}\) and observed engagement are
interpreted descriptively as external comparisons. Consequently, CPS
should be interpreted as a psychologically grounded content-ranking
signal rather than as a direct estimate of platform engagement.

CPS distributions reflect dataset-specific regimes (Table~\ref{tab:cps_logmax}). The Constraint dataset exhibits a skewed upper tail, consistent with evidence that diffusion processes are often dominated by rare, highly amplified outliers \cite{vosoughi2018spread}. COVID--19\_FNIR shows a more bounded CPS range, consistent with recurrent narrative templates and fewer extreme high-risk instances under the same scoring function. Mpox displays a tighter CPS band, including near-zero scores, suggesting that many posts contain weaker combinations of the selected psychology-aligned cues used by the engagement-independent CPS definition and are therefore assessed as having lower relative diffusion propensity under the same scoring function.

To reduce the influence of extreme values and support readable comparison across datasets, $\log_{10}$ of the CPS maximum is reported alongside raw maxima where appropriate (Table~\ref{tab:cps_logmax}). This representation highlights relative extremity and tail behaviour without overstating cross-dataset commensurability of absolute scales. Overall, CPS complements engagement-based indicators by providing interpretable, theory-aligned cues that can support prioritisation and audit when engagement annotations are limited or delayed \cite{slater2007reinforcing}.

\subsection{Comparative Model Performance and Generalisation}
The results reveal trade-offs between classification fidelity, propagation alignment, and behavioural interpretability. On COVID--19\_FNIR, the fusion model achieves near-ceiling classification performance under the current evaluation protocol, which should be interpreted as within-dataset performance rather than guaranteed out-of-domain generalisation. The Constraint dataset also performs strongly for detection, but its diffusion-risk evaluation indicates a tail-dominated regime in which strong detection performance does not necessarily translate to close distributional alignment. Mpox presents the greatest challenge for classification and shows evidence of distribution mismatch despite comparatively strong rank ordering, highlighting dataset dependence and the sensitivity of diffusion-oriented modelling to regime differences and target definitions.

Relative to literature baselines, the proposed model improves on reported results for the Constraint dataset and COVID--19\_FNIR (Table~\ref{tab:baseline_comparison}), but underperforms on the Monkeypox dataset. This pattern is consistent with the possibility that sequential architectures can better capture local dependencies in shorter, syntactically sparse posts \cite{mohbey2024cnn}, and it motivates dataset-aware architectural adaptation, including stronger modelling of short-text structure and additional regularisation for bounded diffusion regimes.

Taken together, these findings suggest that psychologically informed, interpretable feature fusion can strengthen detection and provide a meaningful behavioural lens for analysing diffusion propensity. At the same time, the dataset-specific results emphasise that both linguistic variance and diffusion-risk regime must be considered when interpreting cross-dataset performance and when translating models into operational monitoring settings.

\subsection{Practical \& Theoretical Implications}
The results reinforce the view that health misinformation is not solely a linguistic classification problem, because diffusion risk is modulated by affective framing, norm cues, and heuristic processing in addition to semantic content \cite{lazer2018science,islam2020covid}. Framing the feature design through ELM and TPB provides a principled lens for operationalising these behavioural dimensions \cite{petty1986elaboration,ajzen1991theory}, which is particularly valuable when platform engagement traces are incomplete or unavailable and diffusion-oriented evaluation must rely on text-derived supervision.

From an applied perspective, the findings suggest that intervention strategies should be tailored to the diffusion risk regime and the supervision available for each dataset. In proxy-supervised settings (Constraint and COVID--19\_FNIR), the model output should be interpreted as relative propensity-to-spread rather than engagement-equivalent virality, so robust ranking and prioritisation is typically more defensible than absolute thresholding. In engagement-supervised settings (Monkeypox dataset), distributional mismatch can persist even when rank ordering is strong, which motivates local calibration and pragmatic alerting rules aligned to the bounded spread regime.

\begin{itemize}
    \item \textbf{Constraint dataset (tail-heavy proxy regime):} prioritise outlier-aware rules, for example percentile-based thresholds (top 1\% to 5\%) and log-scaling to stabilise extreme values.
    \item \textbf{Monkeypox dataset (bounded engagement regime):} consider locally tuned filtering and shorter-context modelling, because cue variance and the CPS range can be comparatively constrained.
    \item \textbf{COVID--19\_FNIR (proxy regime with stable ranking):} support periodic recalibration of proxy decision thresholds, especially under shifting narrative composition and rumour saturation.
\end{itemize}

To complement these implications, Table~\ref{tab:guidance} maps observed propagation regimes to practical intervention strategies.

\begin{table*}[t]
\centering
\caption{Guidance for applying CPS-based prioritisation under different diffusion-risk regimes.}
\label{tab:guidance}
\footnotesize
\setlength{\tabcolsep}{4pt}
\renewcommand{\arraystretch}{1.15}
\begin{tabularx}{\textwidth}{@{}>{\raggedright\arraybackslash}X
                        >{\raggedright\arraybackslash}X
                        >{\raggedright\arraybackslash}X@{}}
\toprule
\textbf{Propagation regime} & \textbf{Indicative characteristics} & \textbf{Suggested intervention} \\
\midrule
Tail-heavy (outlier-dominated) &
Highly skewed targets with extreme outliers, often reflected as a heavy upper tail under proxy supervision (for example, Constraint). &
Use percentile thresholds (top-$q$). Apply log-scaling and robust rank-based metrics. Prefer top-$k$ review queues over absolute cut-offs. \\
\addlinespace
Stable ranking under proxy supervision &
Consistent ordering with low error under a fixed protocol (for example, COVID--19\_FNIR under propensity-to-spread targets). &
Use fixed ranking-based triage with scheduled recalibration of thresholds. Prioritise psychologically informed cues when engagement is delayed or missing. \\
\addlinespace
Routinised or repetitive spread &
Recurring rumours and periodic bursts, with topic recirculation over time. &
Recalibrate on a schedule. Track drift in cue distributions, and consider time-aware features and rolling windows. \\
\addlinespace
Bounded engagement with local heterogeneity &
Engagement exists, but exhibits bounded ranges and dataset-specific heterogeneity (for example, Monkeypox dataset). &
Use locally tuned thresholds. Combine CPS with cluster-level monitoring, and apply dataset-specific calibration of alerting rules. \\
\bottomrule
\end{tabularx}
\end{table*}

\subsection{Limitations \& Future Directions}
Several limitations contextualise the scope of these findings and motivate further validation before broader deployment.

First, multilingual and cross-cultural generalisability was not evaluated. The rhetorical and psychological proxies are operationalised using English-centric resources and norms, so transfer to other languages, dialects, and culturally situated persuasion styles remains untested \cite{hofstede2001culture}. Future work should evaluate multilingual variants via domain adaptation, multilingual pretraining, and cross-cultural robustness checks.

Second, CPS is theory-motivated within the broader ELM and TPB framing of this study \cite{petty1986elaboration,ajzen1991theory}, but its post-hoc scoring rule operationalises a narrower set of psychology-aligned textual cues and has not yet been validated through human-centred evaluation. Future studies should test alignment with user perception and expert judgement, for example via share-intent ratings, public health reviewer assessments, and controlled attention or deliberation studies, to examine whether CPS-ranked content aligns with hypothesised central or peripheral cues \cite{martel2020reliance,pennycook2019fighting}.

Third, diffusion-oriented evaluation is sensitive to design choices, including target definitions (observed engagement versus propensity-to-spread proxy), virality thresholds, top-$k$ settings, and partitioning protocols. Future work should run systematic sensitivity analyses across threshold regimes and repeated runs, and should report uncertainty with appropriate resampling-based procedures, for example bootstrap confidence intervals or permutation tests, to quantify the stability of observed differences.

Fourth, the propensity-to-spread targets used for Constraint and COVID--19\_FNIR are deterministic functions of text-derived features, some of which overlap conceptually with information available to the fusion model. Consequently, high agreement with the transferred proxy
should be interpreted as successful learning of the proxy-defined diffusion-risk signal rather than as independent validation of real-world propagation. External validation against observed diffusion traces is required before such scores can be interpreted as engagement-equivalent
estimates.

Fifth, subtle pragmatic phenomena remain challenging, including sarcasm, euphemism, implicit persuasion, and culturally coded phrasing. Addressing these limitations likely requires richer context modelling, conversation-level structure, and targeted annotation schemes, to reduce misclassification of indirect or ironic misinformation \cite{pennycook2019fighting}.

\paragraph{Future work should therefore:}
\begin{enumerate}
    \item Validate CPS using human-centred protocols and expert annotation.
    \item Extend the pipeline to multilingual settings via multilingual or domain-adaptive pretraining.
    \item Add statistical testing and sensitivity analysis over target definitions, thresholds, and repeated runs.
    \item Extend to multimodal misinformation, for example image-text content \cite{singhal2020spotfake+,nakamura2020fakeddit}.
\end{enumerate}

\subsection{Explainability \& interpretability of the fusion model}
\label{subsec:fusion_explainability}
A core motivation for the multi-branch fusion architecture is to support strong performance while enabling transparent, feature-level interpretation for researchers, platform moderators, and public health practitioners. In this study, explainability is primarily achieved through intrinsically interpretable feature design and configuration-level evidence, rather than relying exclusively on post hoc attribution.

The ELM branch separates central-route proxies, such as readability and lexical diversity, from peripheral-route proxies, such as punctuation ratios, capitalisation, and urgency terms. The TPB branch encodes attitude, subjective norms, and perceived behavioural control using proxies including sentiment polarity, group pronouns, social comparison terms, modal verbs, certainty markers, and imperative language. Because these inputs are explicit and theory-aligned, outputs can be interpreted in terms of concrete cue combinations, for example high urgency with strong certainty cues and norm-based framing, rather than only through latent embeddings.

Interpretability is further supported by the explicit separation
of semantic, rhetorical, stance, and psychological streams
(Figure~\ref{fig:fusion_architecture}). The ablation study (Table~\ref{tab:ablation}) provides configuration-level evidence of which branches contribute to performance. The semantic (text) branch drives the largest gains, while the psychological branch yields consistent improvements across datasets, indicating complementary value when affect, norms, or perceived control cues are informative.

Finally, CPS provides a psychologically motivated auxiliary signal by applying the fixed post-hoc scoring rule in Eq.~\eqref{eq:cps-raw} to selected psychology-aligned cues associated with relative diffusion propensity. CPS is engagement-independent by construction because its subfeatures exclude observed engagement fields. Where engagement is available, it is used for propagation-regression supervision, auxiliary evaluation, and proxy validation, but not for CPS construction. CPS summary statistics (Table~\ref{tab:cps_logmax}) and diffusion-oriented results (Table~\ref{tab:propagation}) therefore support interpretation of prioritisation behaviour within the scope of text-derived supervision.

\section{Conclusion}
This study presented a multi-branch fusion model for health misinformation detection and diffusion-oriented analysis, integrating ELM and TPB proxies with rhetorical, stance, and contextual semantic features. Across three pandemic-related datasets, the results show that multi-perspective feature fusion can strengthen within-dataset misinformation detection and can support interpretable analysis of diffusion propensity using CPS.

CPS, as a central contribution, is a theory-aligned auxiliary metric intended to estimate diffusion propensity using psychologically grounded, text-derived proxies. CPS is not positioned as a gold-standard virality measure, but as an interpretable ranking signal that complements engagement-based indicators and remains usable when engagement traces are sparse or delayed.

Empirically, the fusion model improves over reported baselines on the Constraint and COVID--19\_FNIR datasets, while underperforming on the Monkeypox dataset, highlighting dataset dependence and motivating architecture selection based on text sparsity, cue variance, and the diffusion-risk regime. Future work should validate CPS through human-centred studies, extend to multilingual and multimodal domains, and add statistical testing and sensitivity analyses across target definitions and threshold settings.

\bibliographystyle{ieeetr}
\bibliography{refs}
\EOD

\end{document}